\documentclass{article}
\makeatletter
\providecommand{\@trackname}{}
\makeatother

\usepackage[nonatbib, final]{neurips_2026}

\usepackage[utf8]{inputenc} 
\usepackage[T1]{fontenc}    
\usepackage{hyperref}       
\usepackage{url}            
\usepackage{booktabs}       
\usepackage{amsfonts}       
\usepackage{nicefrac}       
\usepackage{microtype}      
\usepackage{xcolor}         

\usepackage{siunitx}

\usepackage{todonotes}
\usepackage{multirow}
\usepackage{makecell}
\usepackage{amsmath}

\usepackage{pifont}
\usepackage{subcaption}
\usepackage{wrapfig}
\usepackage{biblatex}

\newcommand{\xmark}{\ding{55}}
\def\onedot{.}
\newcommand{\ie}{i.e\onedot,}
\newcommand{\eg}{e.g\onedot,}
\newcommand{\etc}{etc\onedot,}

\title{VolCo: Volumetric Contact \\ for High-Fidelity Human Grasp Generation}

\author{%
  Zhuo Chen\textsuperscript{\textnormal{1}}\\
  \texttt{zxc417@student.bham.ac.uk} \\
  \And
  Yihua Cheng\textsuperscript{\textnormal{2, 1}}\thanks{Corresponding author}\\
  \texttt{yihua.cheng@bit.edu.cn} \\
  \And
  Ale\v{s} Leonardis\textsuperscript{\textnormal{1}}\\
  \texttt{a.leonardis@bham.ac.uk} \\
  \And
  Hyung Jin Chang\textsuperscript{\textnormal{1}}\\
  \texttt{h.j.chang@bham.ac.uk} \\
  \Affil
  \textsuperscript{1}University of Birmingham, \textsuperscript{2}Beijing Institute of Technology
}

\begin{document}

\maketitle

\begin{abstract}
    Accurate contact modeling is fundamental to understanding hand–object interaction, yet existing contact representations are typically restricted to object surfaces and rely on hand‑crafted rules to recover contact details, leading to severe penetrations and implausible results. To better exploit the rich detail in motion‑capture data, we introduce \textbf{Vol}umetric \textbf{Co}ntact (VolCo), a representation that expands surface points to a set of 3D volumetric grids. VolCo encodes 3D contact that allows precise hand part recovery, and is organized in an inherent hierarchy: local contact details within each volume and global hand geometry across all volumes. Our framework, VolCoDiff, employs two modules to capture local and global features following this hierarchy. For local contact details, we use a 3D variational autoencoder to model the possible hand configurations conditioned on the local object signed distance field (SDF). For global hand geometry, we design a prior‑guided diffusion model that learns the distribution of compressed latent features aggregated from the volumetric grids. We evaluate our method on two benchmark datasets and demonstrate state‑of‑the‑art performance in penetration and stability, indicating the capability to generate tight grasps with much less severe penetrations. Our code is available at \url{https://github.com/chzh9311/volco}.
\end{abstract}

\section{Introduction}

Hands are the primary means by which we interact with our surroundings. Recently, the research community has dedicated significant attention to modeling object grasping \cite{yangCPFLearningContact2021,zhouTOCHSpatioTemporalObjecttoHand2022,zhangManiDextHandObjectManipulation2025,liLatentHOIGeneralizableHand2025,liuTaskOrientedHumanGrasp2025} due to its broad applications in AR/VR \cite{grady2024Pressurevisionpp}, robotics \cite{Li2024ShapeGraspZT,yang2025egovlalearningvisionlanguageactionmodels}, and human-computer interaction \cite{guo2021human}.

As a fundamental problem in this field, human grasp generation aims to synthesize plausible human hand grasps for a given object template, where MANO \cite{romeroEmbodiedHandsModeling2017} is used as the mainstream hand representation. MANO parameters are predicted directly  \cite{taheriGRABDatasetWholeBody2020,jiangHandObjectContactConsistency2021,zuoGraspDiffGraspingGeneration2024,xuDexterousGraspTransformer2024} or fitted from 3D fields \cite{karunratanakulGraspingFieldLearning2020,yeGHOPGenerativeHandObject2024,moralesVersatileDifferentiableHandObject2024} or contact maps indicating the contact regions, hand part orientations, \etc\cite{zhouTOCHSpatioTemporalObjecttoHand2022,liuContactGenGenerativeContact2023,zhangNL2ContactNaturalLanguage2024,liuTaskOrientedHumanGrasp2025,moralesVersatileDifferentiableHandObject2024, chenForceAware3DContact2025}. 

\begin{figure}[t]
    \centering
    \includegraphics[width=\linewidth]{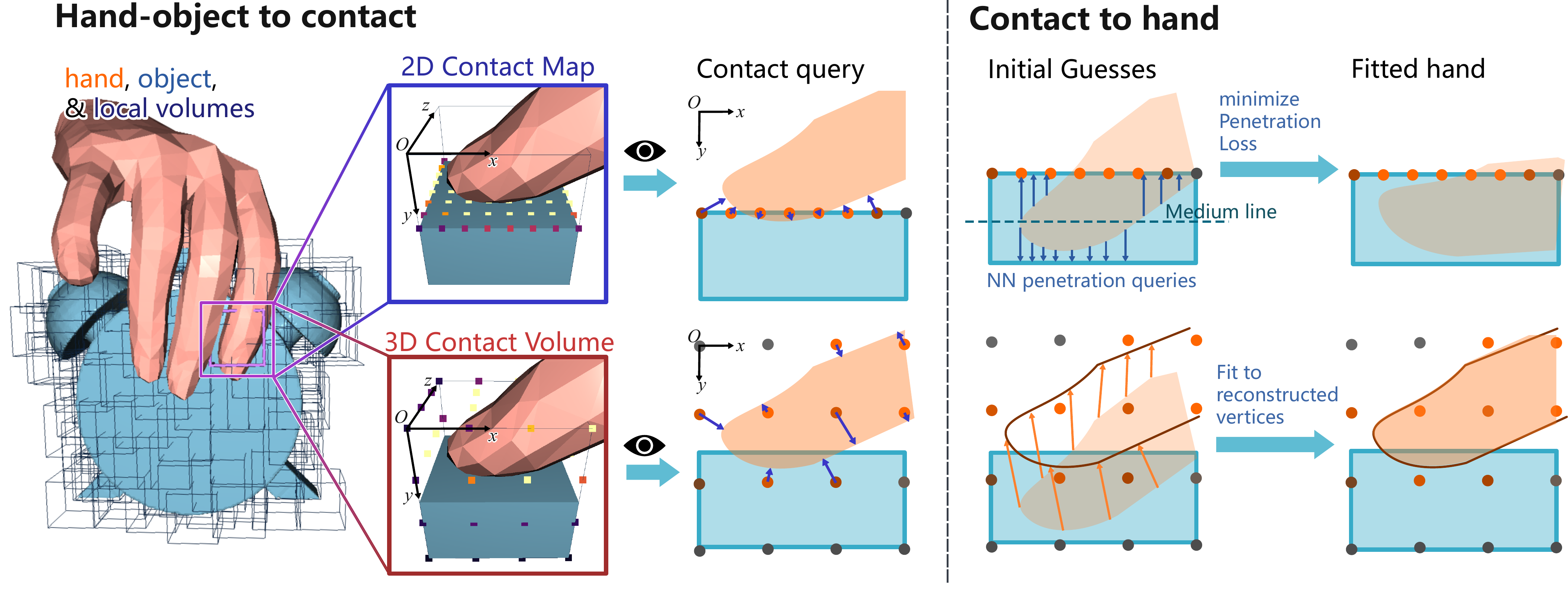}
    \caption{3D contact volumes vs. 2D contact maps. Previous contact maps query points only on object surfaces, which can retain only a small fraction of hand surface. Thus it requires nearest-neighbor (NN) queries to avoid penetrations, which can easily stick in local minima. Contact volumes, however, sample points around the object surfaces and is able to retain the whole local hand configuration in the volume. Consequently, contact volumes can deterministically and precisely reconstruct local hand part and guide the pose fitting without instable penetration losses.}
    \label{fig:teaser}
    \vspace{-1.5em}
\end{figure}

Existing methods mostly rely on hand-crafted rules in test-time-adaptation (TTA) to control penetrations \cite{gradyContactOptOptimizingContact2021}. Such TTA has limited adaptivity since the rules rely on object surface shapes. As shown in Fig.~\ref{fig:teaser}, penetration losses based on nearest-neighbor (NN) queries is non-robust, changing gradient directions when the point crosses the medium line. Moreover, slight penetrations are expected at contact regions with rigid models like MANO, but this method cannot control the exact penetration depth. Consequently, some generated grasps still show severe penetrations and limited plausibility.




Our target is to retain the fine-grained contact details from training data to the largest extent, thus reducing severe penetrations and improving geometric plausibility. Our insight is that the points to reason contact, \ie contact points, should not only lie \textit{on} the object surface, but should be scattered \textit{around} the surface. In this way, the contact can cover much larger area of the hand and preserve the exact penetration depth, thus no rules with NN queries are required. Further, as 3D grid representations \cite{taheriGRABDatasetWholeBody2020,yeGHOPGenerativeHandObject2024,moralesVersatileDifferentiableHandObject2024} are limited by computational cost \cite{liuPointVoxelCNNEfficient2019}, these points should be hierarchically organized to find the balance between computational cost and granularity.





To this end, we introduce \textbf{Vol}umetric \textbf{Co}ntact (VolCo), a novel contact representation constructed from local volumetric grids whose centers are uniformly sampled on the object surface, inspired by Mosaic-SDF~\cite{yarivMosaicSDF3DGenerative2024a}. While Mosaic-SDF employs such local volumes to represent general object geometry via signed distance values, VolCo extends the paradigm to contact modeling of a fixed topology (hand) by capturing also the correspondences. Specifically, each point in a volume encodes the contact likelihood and the correspondence using Continuous Surface Embeddings (CSE) \cite{neverovaContinuousSurfaceEmbeddings2020}. Furthermore, VolCo aggregates point-level features within each local volume. This coherent hierarchy allows the separate modeling of local configurations and global hand priors.
Based on the hierarchy, we present a volume-wise latent diffusion model \ie VolCoDiff, consisting of a 3DVAE \ie VolumeVAE and a latent diffusion model based on Scene Diffuser \cite{huang2023diffusion}.

VolumeVAE aggregates the common prior over local volumes conditioned on the SDF values, modeling \textit{what hand surface configurations are geometrically possible within each volume}. Since each volume is small ($2^3$\unit{\centi\metre^3}), the feasible configuration space is limited in scale and thus possible to be compressed to a compact latent code. VolumeVAE abstracts away the fine-grained surface geometry, allowing the diffusion model to focus solely on the global hand prior.

The latent diffusion model is trained to capture \textit{what combinations of local hand configurations constitute a plausible grasp}. This objective is governed by two key principles. First, all local hand parts must be \textit{reachable} by a valid hand. To enforce this, an auxiliary branch is incorporated in the diffusion UNet to predict the hand pose, which is subsequently mapped to VolCo latent space to supervise the generated contact. Second, the grasp should be \textit{stable}. Due to the joint encoding of SDFs and contact likelihood, VolCo allows the explicit contact force inference via a spring-damper model, and thus a force-aware stability loss can be applied \cite{chenForceAware3DContact2025}. Together, these two principles encourage the generation of physically plausible hand part combinations, while VolumeVAE ensures all local penetrations remain within plausible bounds. Finally, the generated grasp is obtained by initializing with the hand-branch prediction and subsequently fitting it to the generated VolCo representation.

In experiment, we benchmark hand pose reconstruction from contact and compare VolCo with previous representations. VolCo shows the highest precision, indicating its capability to preserve the maximum hand information. We also benchmark the grasp generation capability on two public datasets, and our method achieves the highest stability, lowest penetration depth, and highest plausible contact area. It indicates the state-of-the-art grasping generation quality. In all, our contribution can be summarized in threefold:

\begin{enumerate}
    \item We propose VolCo, a novel volume-based hierarchical contact representation capable of recovering the fine details of hand-object interaction;
    \item We construct a VolCoDiff, a latent diffusion model for grasp generation that takes advantage of the hierarchical structure of VolCo by modeling local contact details using VAE and global hand geometry using diffusion;
    \item We conduct experiments on pose reconstruction from contact and grasp generation, demonstrating state-of-the-art performance on both tasks.
\end{enumerate}

\section{Related Work}

\textbf{Contact Modeling}. Contact modeling has shown great potential in hand-object estimation and generation \cite{jiangHandObjectContactConsistency2021,gradyContactOptOptimizingContact2021}. Early approaches define contact as per-point likelihood on the object surface \cite{gradyContactOptOptimizingContact2021,jiangHandObjectContactConsistency2021,tseS$^2$ContactGraphBasedNetwork2022}, providing guidance on grasp location but leaving a substantial information gap to complete hand meshes. Later, the contact maps get enriched by including part / anchor label \cite{yangCPFLearningContact2021,moralesVersatileDifferentiableHandObject2024,zuoGraspDiffGraspingGeneration2024}, hand continuous surface embedding \cite{neverovaContinuousSurfaceEmbeddings2020,zhangManiDextHandObjectManipulation2025}, part orientations\cite{liuContactGenGenerativeContact2023}, and contact forces \cite{chenForceAware3DContact2025}. Nevertheless, all such representations remain confined to the object surface and lack explicit modeling of penetration depth, leaving its recovery hand-crafted rules.

\textbf{3D Hand-Object Representations}. Some recent methods use implicit functions \cite{karunratanakulGraspingFieldLearning2020,chenGSDFGeometryDrivenSigned2023} or 3D volumetric grids (Bases Point Set \cite{taheriGRABDatasetWholeBody2020,moralesVersatileDifferentiableHandObject2024} or SDF \cite{yeGHOPGenerativeHandObject2024,chenGSDFGeometryDrivenSigned2023}) to represent both hands and objects, thus relating them in a common feature space. Such representations effectively capture relations between hands and objects, but are limited in fitting to known templates or low resolution \cite{liuPointVoxelCNNEfficient2019}. On contrary, VolCo is rooted on known templates and finds the balance of fine-details and computational cost.

\noindent\textbf{Human Grasp Generation}. Human Grasp Generation methods have evolved significantly in recent years. Early methods \cite{karunratanakulSkeletonDrivenNeuralOccupancy2021,liuContactGenGenerativeContact2023,zhengCAMSCAnonicalizedManipulation2023} utilize cVAE \cite{kingmaAutoEncodingVariationalBayes2014} as the backbone, but are limited by the capacity and posterior collapse \cite{bowman2016generating} of VAEs. Subsequent research \cite{shimadaMACSMassConditioned2024,zhangManiDextHandObjectManipulation2025,zuoGraspDiffGraspingGeneration2024,chaText2HOITextguided3D2024,liuTaskOrientedHumanGrasp2025} gradually shifts to diffusion models \cite{hoDenoisingDiffusionProbabilistic2020}, with latent diffusion models \cite{rombachHighResolutionImageSynthesis2022} recently taking dominance. The compressed latent space can reduce the number of parameters \cite{zuoGraspDiffGraspingGeneration2024} and restrict pose plausibility \cite{liLatentHOIGeneralizableHand2025}. Nevertheless, these methods lay their attention on global hand pose, ignoring the contact details.  
Another line of work try to recover the contact first, then fit the hand to the contact maps \cite{yangCPFLearningContact2021,jiangHandObjectContactConsistency2021,zhouTOCHSpatioTemporalObjecttoHand2022,liuContactGenGenerativeContact2023,moralesVersatileDifferentiableHandObject2024,chenForceAware3DContact2025}. However, due to the incomplete and non-deterministic nature of the contact representations, the optimization requires NN based penetration losses \cite{romeroEmbodiedHandsModeling2017} and risks getting stuck in local minima. 


In summary, prior methods typically requires test-time adaptation with non-differentiable penetration losses for better generation quality. We attribute this limitation to the poor information preserving capability of existing contact maps, which VolCo is designed to address.

\section{Methodology}

\begin{figure}[t]
    \centering
    \includegraphics[width=\textwidth]{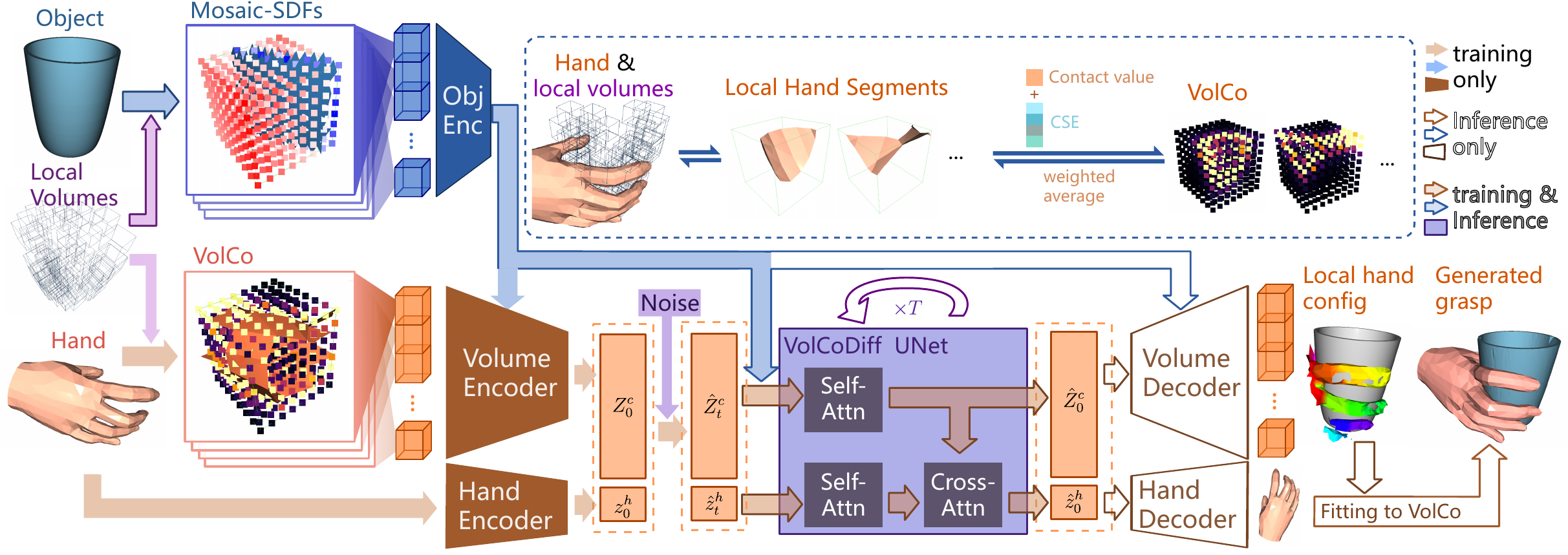}
    \caption{\textbf{VolCo and VolCoDiff framework}. Within the local volumes sampled from the object surface, the hand mesh and VolCo representation are bidirectionally transferable. In each volume, a uniform grid is sampled, and each grid point encodes the contact likelihood and CSE, forming VolCo. Conversely, local hand configurations can be reconstructed from each volume using weighted average and concatenated to recover the hand geometry. VolCoDiff operates on the latent codes extracted from VolCo via VolumeVAE, with an auxiliary hand latent vector for supervision. The generated VolCo restores high-fidelity hand meshes around the object, with the hand decoder output subsequently fitted to VolCo to produce the final grasp.}
    \label{fig:framework}
    \vspace{-1.em}
\end{figure}

Our framework is shown in Fig.~\ref{fig:framework}. VolCo and hand segments within local volumes are mutually convertible. Based on this feature, VolCoDiff is trained to generate the latent and decoded to VolCo, which recovers the hand configurations around the object and gets fitted by MANO.

\subsection{Volumetric Contact Definition}

\noindent\textbf{From Hand-Object to VolCo}. Consider an arbitrary point $\mathbf{x}\in\mathbb{R}^3$ carrying the information of the hand triangular mesh. The contact is represented by $[c_{\mathbf{x}}, \mathbf{e}_\mathbf{x}]$ where $c_\mathbf{x}\in\mathbb{R}$ is the contact likelihood and $\mathbf{e}_\mathbf{x}\in\mathbb{R}^4$ is the 4-D Continuous Surface Embedding (CSE) \cite{neverovaContinuousSurfaceEmbeddings2020} vector of the nearest point on the hand surface. The CSE is obtained following \cite{zhangManiDextHandObjectManipulation2025}. Suppose the nearest point position is $\mathbf{y}$ and the distance is $d$. The vertices of the triangle this point belongs to and the corresponding CSEs are $(\mathbf{v}_i, \mathbf{e}_i), i=1, 2, 3$, then the contact is defined in ContactOpt \cite{gradyContactOptOptimizingContact2021} style and the CSE vector is calculated using the barycentric coordinate to keep continuity as $\mathbf{y}$ moves along the object surface.

\begin{equation}\label{eq:barycentric}
    c_{\mathbf{x}} = \min\{d_0^2/d^2, 1\}; \; \mathbf{e}_{\mathbf{x}}=\begin{bmatrix}\mathbf{e}_1 & \mathbf{e}_2 & \mathbf{e}_3\end{bmatrix}\left(\begin{bmatrix}\mathbf{v}_1 & \mathbf{v}_2 & \mathbf{v}_3\end{bmatrix}^{-1} \mathbf{y}\right).
\end{equation}

\begin{wrapfigure}{r}{.55\textwidth}
    \centering
    \vspace{-1.5em}
    \includegraphics[width=.53\textwidth]{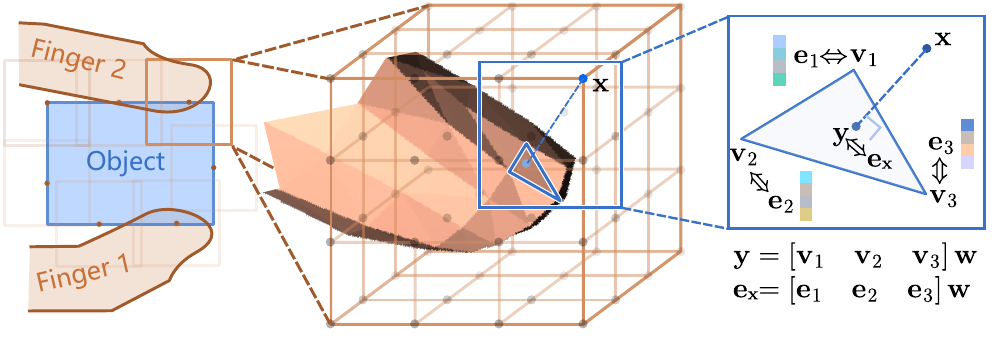}
    \caption{VolCo hierarchy and calculation. Around each of $N$ object surface points, a $k\times k\times k$ volume grid is sampled, where each grid point carries the contact likelihood as a function of $d$ and the CSE of its nearest hand surface point $\mathbf{y}$.}\label{fig:lvc_calc}
    \vspace{-1em}
\end{wrapfigure}

We then discretize the representation and calculate point-wise contact vectors as illustrated in Fig.~\ref{fig:lvc_calc}. Following \cite{yarivMosaicSDF3DGenerative2024a}, we sample $N$ points $\{\mathbf{p}_i\}_{i=1}^N$ from the object surface, and cover the object surface by expanding each point to a local volume with half size $s$. Then the volume is discretized with a resolution of $k$. The volumetric grid points are defined by $\mathcal{G}_i = \left\{\mathbf{x} =\mathbf{p}_i + s\cdot (2\mathbf{j} - k - 1 )/(k-1)\right\}_{\mathbf{j} = (i_1, i_2, i_3)\in [1, k]^3}$
    
Define the unit distance as half the grid gap $d_0 = s/(k-1)$, and VolCo for this grid can be written as


\begin{equation} \label{eq:contact_grid}
    C_i=\{[c_\mathbf{x}, \mathbf{e}_{\mathbf{x}}]\}_{\mathbf{x}\in \mathcal{G}_i}.
\end{equation}

\noindent\textbf{From VolCo to Hand}. One key advantage of VolCo the capability to reconstruct the hand part within the volume. For each point contact $(\hat{c}_{\mathbf{j}}, \hat{\mathbf{e}}_{\mathbf{j}})\in G_i^h$, we find the triangle $\mathbf{t}_{\mathbf{j}}$ on the mesh with the nearest CSE (defined by the average CSE of 3 vertices) to $\hat{\mathbf{e}}$. Then, according to Eq.~\eqref{eq:barycentric}, the corresponding weight of the nearest point to the 3 vertices can be equivalently calculated by  $\mathbf{w}_\mathbf{j}=\begin{bmatrix}\mathbf{e}_1 & \mathbf{e}_2 & \mathbf{e}_3\end{bmatrix}^{-1}\hat{\mathbf{e}}_{\mathbf{j}}\in\mathbb{R}^3$. The weight vector can then be extended to $\bar{\mathbf{w}}_{\mathbf{j}}\in\mathbb{R}^{778}$ covering all vertices by padding the rest indices with $0$. For a specific vertex $h$ that appears at least once in the nearest triangle query process, \ie $\mathbf{v}_h\in\cup_{\mathbf{j}\in[1,k]^3} \mathbf{t}_{\mathbf{j}}$, the position can be deterministically estimated as the weighted average of all volume points using

\begin{equation}\label{eq:recon}
    \hat{\mathbf{v}}_h = \frac{\sum_{\mathbf{j}\in[1,k]^3}\hat{c}_\mathbf{j}\bar{w}_{\mathbf{j}h}\mathbf{x}_\mathbf{j}}{\sum_{\mathbf{j}\in[1,k]^3}\hat{c}_\mathbf{j}\bar{w}_{\mathbf{j}h} + \varepsilon},\quad \varepsilon = 10^{-8}.
\end{equation}
In our experiment, we set $k=8$ by default, while $k=4$ setting is also reported to study the influence of resolution on efficiency and performance. $s$ is fixed to 1\unit{\centi\metre} since natural hand deformations cannot exceed 1\unit{\centi\metre}. Fixed scale also facilitates capturing consistent hand local configurations regardless of the object scale. Empirically, $N=128$ volumes roughly cover all objects in YCB benchmark \cite{chaoDexYCBBenchmarkCapturing2021}.



\subsection{Modeling Contact Prior using VolumeVAE}

The hierarchical structure of VolCo allows us to decouple the modeling of local contact details and global geometry. This section mitigates the local hierarchy, answering the question: \textbf{How to model the distribution of feasible local hand configurations given the SDF values?} 

Since local hand configurations are limited in shape variations and only occupy a part of the contact feature space, we propose to use a 3D VAE, \ie VolumeVAE, following the practice in 3D U-Net \cite{cicek3DUNetLearning2016} and 3DVAEs \cite{brockGenerativeDiscriminativeVoxel2016}, to compress the contact volume $C\in\mathbb{R}^{k\times k\times k \times (1+4)}$ into a compact 128-dimensional latent vector conditioned on the SDF values at each point $D=\{\text{SDF}_o(\mathbf{x})\}_{\mathbf{x}\in\mathcal{G}}$.



 
\textbf{Losses}. Formally, VolumeVAE models the conditional distribution $p(C \mid D)$ via a standard VAE formulation with reparameterization, where the encoder outputs $(\mu, \sigma)$ and the decoder outputs $\hat{C}$ from the sampled latent. Generally, VAE supervision is done by reconstruction loss $\mathcal{L}_{rec}$ and KL-Divergence loss $\mathcal{L}_{KL}$. In our case, $\mathcal{L}_{KL} = D_{\text{KL}}\left(q_\theta(z \mid C, D) \,\|\, p(z)\right)$ where $p(z) = \mathcal{N}(0, 1)$. However, the common reconstruction loss cannot precisely reconstruct the part hand surface because the contact and correspondences need to be treated differently.

For contact, a simple MSE loss is enough: $\mathcal{L}_{c} = \|c_i - \hat{c}_i\|_2$. According to Eq.~\eqref{eq:recon}, we also regularize the CSE loss using contact values: $\mathcal{L}_{cse} = c_i\|\mathbf{e}_i - \hat{\mathbf{e}}_i\|_1$. Ideally, a reconstruction loss of hand vertices should be applied, yet in practice the nearest triangular face lookup is not differentiable. Therefore, we propose the \textit{CSE weight loss}, which supervises the 3-dimensional barycentric weights based on ground truth triangle. Conceptually, we want $\hat{\mathbf{w}} =  \begin{bmatrix}\mathbf{v}_1 & \mathbf{v}_2 & \mathbf{v}_3\end{bmatrix}^{-1} \hat{\mathbf{y}} $. Due to the differentiability issue, we equivalently have $\hat{\mathbf{w}} = \begin{bmatrix} \mathbf{e}_1 & \mathbf{e}_2 & \mathbf{e}_3 \end{bmatrix}^{-1} \hat{\mathbf{e}}$ according to  Eq.~\eqref{eq:barycentric}, leading to a continuous loss $\mathcal{L}_{csew} = \|\hat{\mathbf{w}}- \mathbf{w}\|_1$. The final loss for training the VolumeVAE is:

\begin{equation}\label{eq:gridae_loss}
    \mathcal{L}_{vae} = \lambda_{c}\mathcal{L}_{c} + \lambda_{cse}\mathcal{L}_{cse}  + \lambda_{csew}\mathcal{L}_{csew} + \beta \mathcal{L}_{KL}.
\end{equation}
In experiments, we set $\lambda_{c} = 5.0, \lambda_{cse}=1.0, \lambda_{csew}=0.5$. $\beta$ is set to $10^{-6}$ following~\cite{rombachHighResolutionImageSynthesis2022}.

\subsection{Prior-Guided Latent Diffusion}

VolumeVAE encodes rich information about possible local hand configurations, so the next step is to answer: \textbf{How to model the distribution of possible combinations of local hand configurations?}

Latent diffusion models \cite{rombachHighResolutionImageSynthesis2022} has recently shown great potential in human grasp generation \cite{wuFastGraspEfficientGrasp2024,zuoGraspDiffGraspingGeneration2024,liLatentHOIGeneralizableHand2025}. We follow this practice, but with one key difference: the latent space is the concatenation of the volume-wise latent code, following the global hierarchy of VolCo. 

To minimize the range of latent vectors, we set all volumes without hand vertices inside to zero, which are mapped to a constant latent. The resulting latent is a matrix $Z^c \in \mathbb{R}^{N\times 128}$. This volume-wise latent space integrates naturally with Scene Diffuser~\cite{huang2023diffusion}, a point-conditioned diffusion framework. Inspired by \cite{liuTaskOrientedHumanGrasp2025}, we use the flattened Mosaic-SDF as the feature vector and utilize the PointNet~\cite{qi2017pointnet} encoder to extract the per-point features. These features are used as the positional encoding in the attention-based UNet.

\noindent\textbf{Auxiliary Branch with Hand Prior Guidance}. Generated VolCo should be valid, \ie whether they can fit in a valid hand mesh (provided by MANO~\cite{romeroEmbodiedHandsModeling2017}). Since the VolCo carries complete information to recover the hand, an auxiliary branch to generate hand based on VolCo latents is added. Also, to make sure the generated hand fall in the plausible manifold, the target is a latent $z^h\in\mathbb{R}^{16}$ extracted from an MLP-based VAE model. The final latent space is obtained by concatenation: $z = [z^h, Z^c]$.



In general, diffusion models a Markov noising process $\{z_t\}_{t=1}^N$ and $z_0$ is the clean sample. Using object SDFs $D$ as the condition, the reverse denoising process is to extract $p(z_0|D)$ by gradually cleaning $z_t$. We construct the standard diffusion loss following \cite{tevetHumanMotionDiffusion2022} by predicting the clean latent $\hat{z}_0 = f_{\phi}(z_t, t, D)$ and supervise by the standard loss $\mathcal{L}_{diff} = \|z_0-f_\phi(z_t, t, D)\|^2_2$.



To allow the hand prior to flow into VolCo latent generation, we develop a single-directional dataflow. The features from the object and VolCo latents serve as the condition and fed into the hand generation branch by cross-attention \cite{vaswaniAttentionAllYou2017}. With the predicted hand mesh $\langle \hat{V}, \mathcal{F}\rangle$, we can reconstruct the contact via Eq.~\eqref{eq:contact_grid} and then the latent using VolumeVAE decoder. Then the difference loss is calculated by comparing the reconstructed latent $\hat{Z}^c = \{\mu_i + \epsilon \sigma_i\}_{i=1}^{k^3}$ with the predicted $\hat{Z}_0^c$ by $\mathcal{L}_{cons} = \|\hat{Z}^c - \hat{Z}_0^c\|_2^2$. The predicted hand mesh is also supervised by reconstruction loss $\mathcal{L}_{rec} = \|\hat{V}_0 - V_0\|_2^2$. Since the nearest point lookup is non-differentiable, the latent reconstruction does not carry gradient. The reconstructed hand supervises the generated contact but not vice versa.

\noindent\textbf{Stability Loss Using Contact Forces}. Another advantage of VolCo is that the contact force can be explicitly inferred. Suppose $k$ is the object stiffness and  $\delta$ is the penetration depth. According to the linear spring-damper model~\cite{hunt1975coefficient} for static contact ($\dot{\delta} = 0$): $F = k \delta$. We can explicitly calculate the contact force for each point, as each point of VolCo encodes both the contact value $c$ and the penetration depth as object SDF. Regard two points as adjacent if their distance is smaller than half the grid gap. Assume $\mathcal{A}_{\mathbf{x}}$ is the adjacent point set of point $\mathbf{x}\in\mathcal{G}_i$. We also add a margin $m=1\unit{\milli\metre}$ and slack variables $\Delta d_{\mathbf{x}}\in[-1\unit{\milli\metre}, 1\unit{\milli\metre}]$ to allow error in penetrations. The force at this point is

\begin{equation}\label{eq:force}
    F(\mathbf{x}) = 1/(|\mathcal{A}_{\mathbf{x}}| + 1)kc_{\mathbf{x}}\cdot \min\{\text{SDF}_o(\mathbf{x}) - m + \Delta d_\mathbf{x}, 0\} \cdot \nabla \text{SDF}_o(\mathbf{x}),
\end{equation}
where $1/(|\mathcal{A}_\mathbf{x}|+1)$ averages the force to the adjacent points, $c_{\mathbf{x}}$ regularizes the force value using contact, $\nabla{\text{SDF}}_o(\mathbf{x})$ points the force towards the SDF descending direction. The stability loss $\mathcal{L}_{stability}$ is improved from \cite{chenForceAware3DContact2025} by adding the slack variables. We refer the readers to Appendix.~\ref{app:stability_loss} for detailed formulation.

Finally, with the weights $\lambda_{cons}=\lambda_{rec}=\lambda_{stability}=0.1$ in experiments, the loss in training the diffusion model is written as

\begin{equation}\label{eq:all_diff_loss}
    \mathcal{L}_{ldm} = \mathcal{L}_{diff} + \lambda_{cons}\mathcal{L}_{cons} + \lambda_{rec}\mathcal{L}_{rec} + \lambda_{stability}\mathcal{L}_{stability}.
\end{equation}

\subsection{Finetuning the Local Details.}

Although the hand branch reconstructs a feasible hand, it usually does not align exactly with generated VolCo. Thus, finetuning is usually necessary for high-quality output. Different from literature \cite{liuContactGenGenerativeContact2023,yeGHOPGenerativeHandObject2024,chenForceAware3DContact2025}, our goal is nothing but finding the best fit to VolCo near the initial pose provided by hand branch. 

From the predicted VolCo, we can obtain correspondence weights $\hat{\bar{w}}_{\mathbf{x}h}$ between grid point $\mathbf{x}$ and vertex $h$ in Eq.~\eqref{eq:recon}, and the contact likelihood $\hat{c}_\mathbf{x}$. Suppose $M$ volumes are predicted as in contact (not all-zero), then the reconstruction loss is defined based on the weighted sum of MSE between each grid point and its predicted nearest point on the hand mesh.

\begin{equation}
\mathcal{L}_{vrec}= \frac{1}{Mk^3}\sum\nolimits_{i=1}^M\sum\nolimits_{\mathbf{x}\in\mathcal{G}_i} \hat{c}_\mathbf{x}\text{MSE}\left(\mathbf{x}, \sum\nolimits_{h=1}^{778}\hat{\bar{w}}_{\mathbf{x}h}\hat{\mathbf{v}}_h\right).
\end{equation}

Second, for $N-M$ volumes that are predicted as non-contact, a simple repulsive loss is applied to fit those non-contact volumes. For any $\mathbf{x}\in\mathcal{G}_i$, estimate the contact using its nearest hand vertex as $\tilde{c}_{\mathbf{x}}$. Then the repulsive loss is $\mathcal{L}_{rep}=-\sum_{i=1}^M\sum_{\mathbf{x}\in\mathcal{G}_i}\tilde{c}_\mathbf{x} \min\{SDF_o(\mathbf{x}), 0\} / (Nk^3-Mk^3)$.

With the pose from the auxiliary branch as the initial state, a regularization term is applied to punish large modifications: $\mathcal{L}_{reg}=\|\hat{\theta}-\theta_0\|$. Our final loss function for the optimization process is 

\begin{equation}
    \mathcal{L}_{optim} = \mathcal{L}_{vrec} + \lambda_{rep}\mathcal{L}_{rep} + \lambda_{reg}\mathcal{L}_{reg}.
\end{equation}

\section{Experiment}

In this section, we conduct experiments on 2 tasks: reconstructing from ground truth contact and human grasp generation, both done on two public datasets.


\noindent\textbf{Datasets}. GRAB \cite{taheriGRABDatasetWholeBody2020} is used as our training set and one of the test set due to its high quality motion capture data. HO3D \cite{hampaliHOnnotateMethod3D2020} contains 10 objects from YCB dataset \cite{chaoDexYCBBenchmarkCapturing2021} and are not seen when training on GRAB. This dataset is used for benchmarking the adaptivity to out-of-domain objects.

\noindent\textbf{Implementation Details}. VolumeVAE and VolCoDiff are both trained using AdamW \cite{loshchilov2017decoupled} with learning rates $2\times10^{-4}$ and $10^{-4}$ for 20 and 100 epochs, respectively. Pose fitting uses AdamW for 1,000 iterations. Our experiments are done on a linux workstation with Intel(R) Xeon(R) w7-3445 CPU, 64GB RAM, and two 4090 GPUs. Please refer to Appendices~\ref{app:model} and~\ref{app:preprocess} for model architectures and data pre-processing details.



\subsection{Reconstruction from VolCo}



We test the correspondence and contact representations together with comparisons to previous representations on GRAB test set every 64 frames. Three criteria are utilized following previous research \cite{zimmermannFreiHANDDatasetMarkerless2019,hampaliHOnnotateMethod3D2020,liuContactGenGenerativeContact2023}: 1) \textit{End Point Error (EPE)}---average Euclidean distance of mesh vertices; 2) \textit{F-score}---harmonic average of precision and recall between vertices of two meshes under certain threshold(5\unit{\milli\metre}  and 15\unit{\milli\metre}); 3) \textit{Area Under the ROC Curve (AUC)} \cite{FAWCETT2006861} expanding from 0 to 5\unit{\centi\metre}. All hands are aligned to the objects. The resulting poses are obtained via optimization, initialized by the best 6D poses that fit the average hand to the contact. See Appendix~\ref{app:init} for initialization details.

\begin{table}[t]
\caption{Quantitative results of hand reconstruction from ground truth contact on GRAB test set.  `$\uparrow$' after a criterion means the higher the better, while `$\downarrow$' means the opposite. 2D contact maps use the same number of sampled points (4096) as our sparse setting, yet our contact still demonstrates much better performance across all criteria.}\label{tab:gt_contact_recon}
\begin{tabular*}{\textwidth}{@{\extracolsep{\fill}}ccc|cccc@{}}
\toprule
Method & Corr & Contact & EPE({\unit{\milli\metre}}) $\downarrow$  & F@5{\unit{\milli\metre}} $\uparrow$ & F@15{\unit{\milli\metre}} $\uparrow$ & AUC $\uparrow$ \\
\midrule
ContactOpt \cite{gradyContactOptOptimizingContact2021}   &  --     & Map  &   89.42   &  0.125  &  0.365 & 0.067 \\
ContactGen \cite{liuContactGenGenerativeContact2023}       &     Part           &     Map + D           &          59.38            &    0.405                                  &     0.773       &    0.619        \\
ManiDext \cite{zhangManiDextHandObjectManipulation2025}       &     CSE           &     Map            &   11.19   &      0.616             &       0.923                     &      0.785     \\\midrule
 Sparse ($ 64 \times 4^3$)        &   CSE    & Volume    &         6.70\textsuperscript{\tiny\textcolor{red}{$\downarrow$40.1\%}}               &          0.739\textsuperscript{\tiny\textcolor{red}{$\uparrow$20.0\%}}          &    0.970\textsuperscript{\tiny\textcolor{red}{$\uparrow$5.1\%}}   &  0.868\textsuperscript{\tiny\textcolor{red}{$\uparrow$10.6\%}}\\
 Normal ($ 128 \times 8^3$)        &   CSE    & Volume    &         \textbf{6.33}\textsuperscript{\tiny\textcolor{red}{$\downarrow$43.4\%}}              &          \textbf{0.752}\textsuperscript{\tiny\textcolor{red}{$\uparrow$22.1\%}}          &    \textbf{0.971}\textsuperscript{\tiny\textcolor{red}{$\uparrow$5.2\%}}   &  \textbf{0.875}\textsuperscript{\tiny\textcolor{red}{$\uparrow$11.4\%}} \\
\bottomrule
\end{tabular*}
\end{table}

\begin{figure}[t]
    \centering
    \includegraphics[width=.9\textwidth]{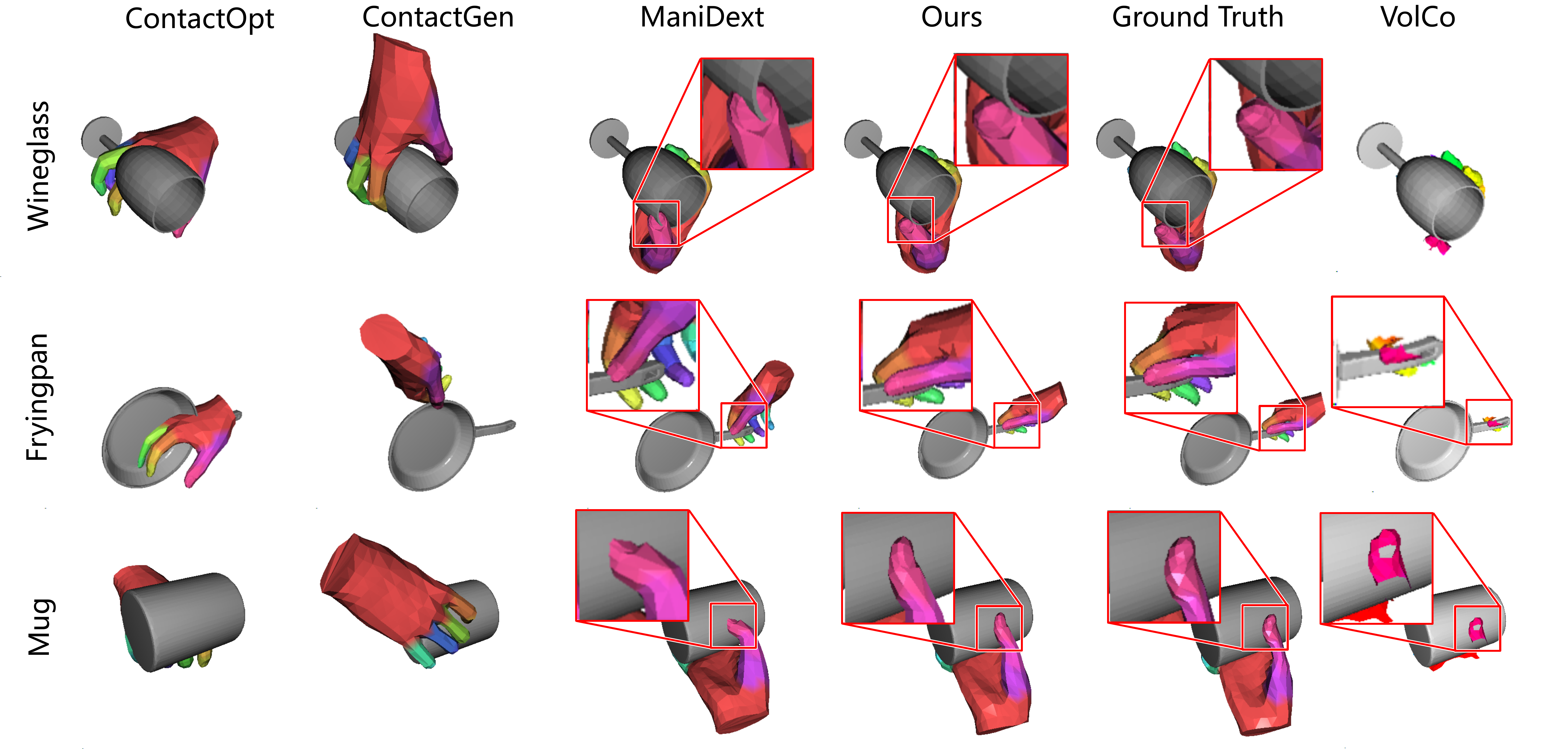}
    \caption{Reconstruction from previous contact maps vs. VolCo. The rightmost column represents VolCo using the reconstructed local vertices. VolCo demonstrates the best capability to restore details.}
    \label{fig:qualitative_local}
    \vspace{-1em}
\end{figure}

In Tab.~\ref{tab:gt_contact_recon}, our method achieves the best performance across all metrics, indicating superior information preservation. The improvement from part labels to CSE confirms that point-wise correspondence is critical for reconstruction. The transition from map to volume representation reduces EPE by 40.1\% at the same point count, highlighting the importance of sampling along the surface normal direction.

Fig.~\ref{fig:qualitative_local} illustrates the reason for improved precision. First, VolCo itself can recover local hand details without MANO as shown in the last column. This makes the fitted model precise at local details, restoring the slight penetrations; Second, as shown in `Fryingpan' example, such contact encodes the exact spatial positions of more points with higher confidence around the object surface, which makes it possible to recover a tight grip around a slim bar---where the contact area is very small.

\subsection{Grasp Generation Analysis}

\textbf{Metrics}. Following previous research \cite{jiangHandObjectContactConsistency2021,liuContactGenGenerativeContact2023,wuFastGraspEfficientGrasp2024,chenForceAware3DContact2025}, we utilize multiple criteria to evaluate stability, geometric plausibility and diversity of the generation. 1) \textit{Simulation Displacement (SD)} measures the displacement of object center of mass in PyBullet \cite{coumans2021} simulation; 2) \textit{Penetration Depth (PD)} measures the average of all samples' deepest penetration depth; 3) \textit{Intersection Volume (IV)} is calculated by voxelizing the object meshes to \SI{1}{m\metre^3} and counting the overlapping voxels. 3) \textit{Contact Ratio (CR)} reflects the ratio of samples in contact. To indicate the tightness of grasps, we introduce \textit{Contact Area (CA)} to measure the valid contact area on the object surface (See Appendix.~\ref{app:contact_area}). 4) \textit{Entropy} and \textit{Cluster Size} measure the diversity of generation.

\textbf{Quantitative Analysis}. The results on both datasets are reported in Tab.~\ref{tab:quant_result_grab} and~\ref{tab:quant_result_ho3d} along with 1-sigma error bars. On the in-domain GRAB dataset, our method achieves the lowest SD and PD with the largest CA, exceeding the previous state-of-the-art by 14.8\%, 32.5\%, and 1.6\%, respectively. This demonstrates that the generated grasps are stable, plausible, and contact-rich — whereas previous methods often collapse to similar poses in exchange for less penetration. On the out-of-domain HO3D dataset, our method surpasses the previous state-of-the-art by 12.3\% (SD), 28.5\% (PD), 18.2\% (CA), demonstrating strong generalization to unseen objects. Despite HO3D objects being generally larger than those in GRAB, our method retains strong adaptability to these scales.

\begin{figure}[t]
    \centering
    \includegraphics[width=.9\textwidth]{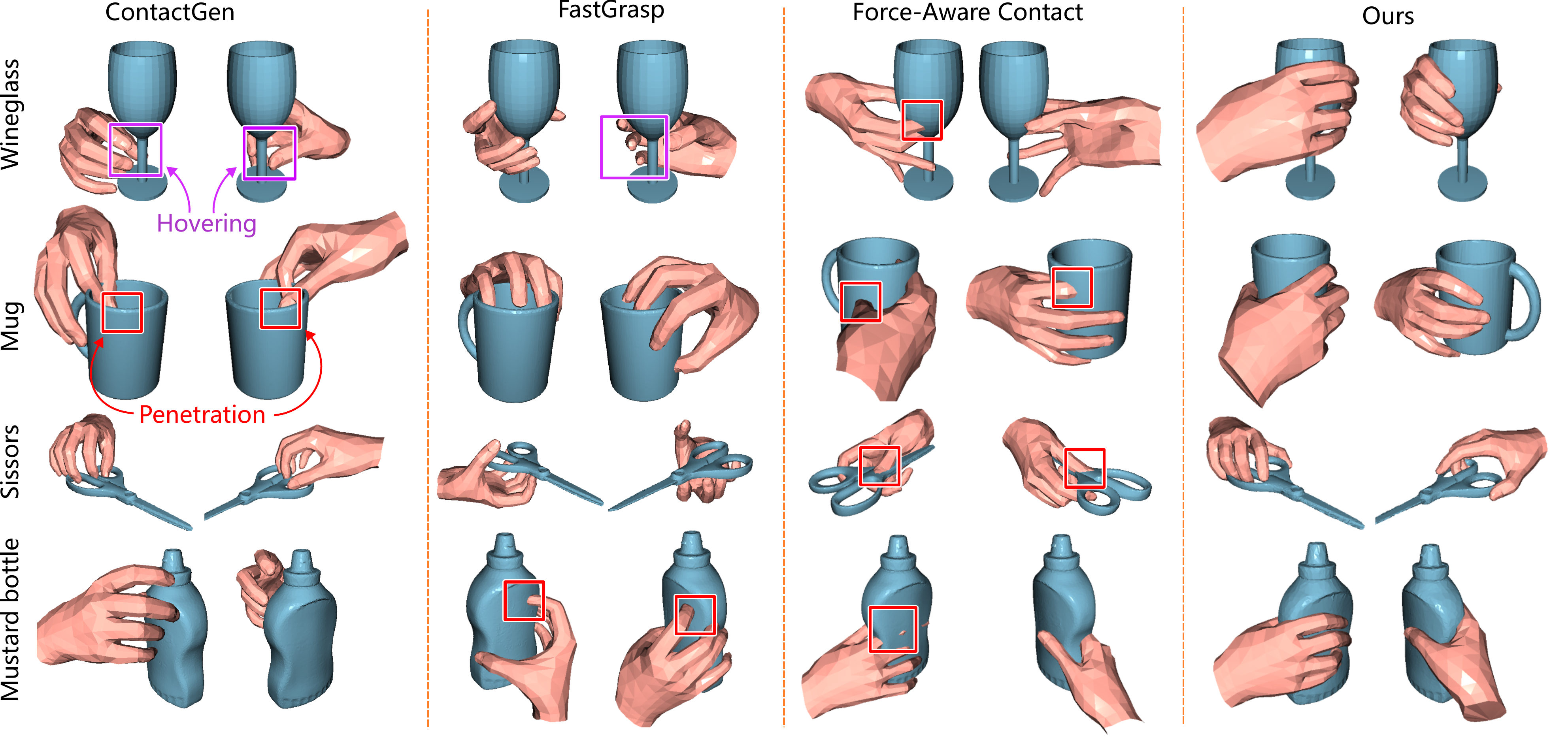}
    \caption{Qualitative comparison with previous methods \cite{liuContactGenGenerativeContact2023,wuFastGraspEfficientGrasp2024,chenForceAware3DContact2025}. Hovering fingers are marked with purple boxes while penetrations are marked in red. While previous methods tend to produce excessively loose or penetrating grasps, our method achieves a physically plausible balance.}
    \vspace{-1em}
    \label{fig:qualitative_global}
\end{figure}

Our intersection volumes are not the least among all methods, which is desired as a result of tighter grasps. Larger CA indicate that our method generate grasps with more valid contact. Since slight penetrations are allowed, larger CA is usually accompanied with larger IV. Over 28\% improvement in PD on both datasets indicates a larger portion of penetrations are plausible. The preference for tighter grasps is also shown in Qualitative analysis in Fig.~\ref{fig:qualitative_global}.

\begin{table}[t]
\centering
\caption{Quantitative results of grasp synthesis on GRAB Dataset. `$\uparrow$' after a criterion means the higher the better, while `$\downarrow$' means the opposite. Best result in each column is marked in \textbf{bold} and the second best is \underline{underlined}. Our method achieves the best stability, smallest PD and largest CA.}\label{tab:quant_result_grab}
\begin{tabular*}{\textwidth}{@{\extracolsep{\fill}}lccccccc@{}}
\toprule
\textbf{Method} & \thead{SD (\unit{\centi\metre})$\downarrow$} & \thead{PD (\unit{\centi\metre})$\downarrow$} & \thead{IV (\unit{\centi\metre^3})$\downarrow$} & \thead{CA (\unit{\centi\metre^2})} $\uparrow$ & \thead{CR $\uparrow$} & \thead{Entropy} $\uparrow$ & \thead{Cluster \\ Size $\uparrow$} \\ \midrule
GrabNet \cite{taheriGRABDatasetWholeBody2020}   & $1.02_{\pm 2.07}$       &   $\underline{0.40}_{\pm 0.34}$ & $7.43_{\pm 9.89}$ & $25.5_{\pm 13.5}$ & \textbf{1.00}           & \textbf{2.79}          & 2.57           \\
ContactGen \cite{liuContactGenGenerativeContact2023} & $2.32_{\pm 3.47}$ & $0.52_{\pm 0.49}$ & $\underline{3.32}_{\pm 3.46}$ & $19.9_{\pm 13.3}$   & \underline{0.99} & 2.76          & \textbf{4.28}          \\ 
FastGrasp~\cite{wuFastGraspEfficientGrasp2024} & $2.55_{\pm 3.71}$ & $0.53_{\pm 0.66}$ & $\textbf{2.02}_{\pm 2.19}$ & $13.3_{\pm 7.42}$     & 0.86   & \underline{2.77}          & 0.41          \\ 
FAGrasp~\cite{chenForceAware3DContact2025}   & $\underline{0.61}_{\pm 1.37}$ & $0.73_{\pm 0.67}$ & $4.74_{\pm 4.71}$ & $\underline{30.6}_{\pm 11.1}$  & \textbf{1.00}          & 2.65          & \underline{3.84} \\ \midrule
Ours & $\textbf{0.52}^\text{\textcolor{red}{$\downarrow$14.8\%}}_{\pm 0.64}$ & $\textbf{0.27}^\text{\textcolor{red}{$\downarrow$32.5\%}}_{\pm 0.28}$ & $3.78_{\pm 4.05}$ & $\textbf{31.1}^\text{\textcolor{red}{$\uparrow$1.6\%}}_{\pm 12.9}$ & \textbf{1.00} & \underline{2.77} & 3.61 \\
\bottomrule
\end{tabular*}
\end{table}


\begin{table}[t]
\centering
\caption{Quantitative results of grasp synthesis on HO3D Dataset. `$\uparrow$' after a criterion means the higher the better, while `$\downarrow$' means the opposite. Best result in each column is marked in bold. Our method shows strong adaptivity to out-of-domain objects.}\label{tab:quant_result_ho3d}
\begin{tabular*}{\textwidth}{@{\extracolsep{\fill}}lccccccc@{}}
\toprule
\textbf{Method} & \thead{SD (\unit{\centi\metre})$\downarrow$} & \thead{PD (\unit{\centi\metre})$\downarrow$} & \thead{IV (\unit{\centi\metre^3})$\downarrow$} & \thead{CA (\unit{\centi\metre^2})$\uparrow$} & \thead{CR $\uparrow$}  & \thead{Entropy} $\uparrow$ & \thead{Cluster\\ Size $\uparrow$} \\ \midrule
GrabNet \cite{taheriGRABDatasetWholeBody2020}  & $1.61_{\pm 2.84}$ & $1.04_{\pm 0.78}$ & $8.30_{\pm 6.69}$ & $31.4_{\pm 17.7}$ & \textbf{1.00}           & \textbf{2.87 }         & 3.18           \\
GraspTTA \cite{jiangHandObjectContactConsistency2021} & $4.17_{\pm 4.21}$ & $1.12_{\pm 0.92}$ & $\textbf{3.40}_{\pm 3.38}$ & $16.6_{\pm 15.5}$ & 0.72 & 2.76 & 0.22 \\
ContactGen \cite{liuContactGenGenerativeContact2023} & $3.41_{\pm 4.31}$ & $\underline{0.98}_{\pm 0.78}$ & $\underline{4.01}_{\pm 2.93}$ & $19.9_{\pm 12.3}$ & \underline{0.95} &    2.77          & \textbf{4.56}          \\ 
FastGrasp~\cite{wuFastGraspEfficientGrasp2024}  & $2.49_{\pm 3.35}$ & $1.18_{\pm 0.88}$ & $9.54_{\pm 19.7}$ & $18.1_{\pm 15.2}$   & \textbf{1.00}   & 2.76          & 0.34          \\ 
FAGrasp~\cite{chenForceAware3DContact2025}     & $\underline{1.21}_{\pm 1.93}$ & $1.93_{\pm 1.33}$ & $45.4_{\pm 93.6}$ & $\underline{38.4}_{\pm 21.8}$ & \textbf{1.00}          & 2.76          & \underline{4.17} \\ \midrule
Ours & $\textbf{1.06}^\text{\textcolor{red}{$\downarrow$12.3\%}}_{\pm 1.68}$ & $\textbf{0.70}^\text{\textcolor{red}{$\downarrow$28.6\%}}_{\pm 0.64}$ & $10.9_{\pm 27.3}$ & $\textbf{45.4}^\text{\textcolor{red}{$\uparrow$18.2\%}}_{\pm 29.2}$ & \textbf{1.00} & \underline{2.85} & 3.85 \\
\bottomrule
\end{tabular*}
\end{table}

\textbf{Qualitative Analysis}. We compare our method qualitatively with previous state-of-the-art methods in penetrations \cite{liuContactGenGenerativeContact2023,wuFastGraspEfficientGrasp2024} and stability \cite{chenForceAware3DContact2025} in Fig.~\ref{fig:qualitative_global}. The figure highlights the contact details, where previous methods tend to generate loose grasps due to aggressive penetration losses or implausibly deep penetrations especially in thin objects like wineglass and mug. This is due to the non-differentiable penetration loss with NN query. Our method, on contrast, finds a good balance of grasping steadiness and penetration---penetrations are common but mostly not severe, just like in the original motion capture data, indicating higher geometric plausibility. This also explains the lower penetration depth and higher penetration volumes, as our model prefers tight grasps with larger contact areas. More generated grasps are available in Appendix~\ref{app:exp_result}.


\subsection{Ablation Study}

This section studies the contribution of VolCo and the key design choices, \ie hand prior guidance and explicit force-aware stability loss. `Prior Guidance` indicates if the auxiliary branch is used during training, while `Initialization' means if the pose is initialized with the predicted pose. The first experiment use point-based contact maps following \cite{liuTaskOrientedHumanGrasp2025}. 

\begin{table}[t]
\centering
\caption{Ablation study on key design choices. The best are marked in \textbf{bold} and the second best are \underline{underlined}. VolCo brings obvious improvement, and the stability loss is more effective in reducing SD with geometric plausibility ensured by prior guidance and initialization.}\label{tab:ablation}
\begin{tabular*}{\textwidth}{@{\extracolsep{\fill}}c|cccc|ccccc@{}}
\toprule
No. & \thead{VolCo} & \thead{Prior \\ Guidance} & \thead{Stability \\Loss}  & \thead{Initia- \\lization} &
\thead{SD \\ (\unit{\centi\metre})} $\downarrow$ & \thead{PD \\ (\unit{\centi\metre})}$\downarrow$ & \thead{IV \\ (\unit{\centi\metre^3})}$\downarrow$ & \thead{CR}$\uparrow$ & \thead{CA}$\uparrow$\\
\midrule
1 & \xmark     & \xmark      & \xmark    & \xmark    & 1.72 & 0.42 & 4.64 & 0.92 & 24.7 \\
2 & \checkmark & \xmark      & \xmark    & \xmark    & 1.28 & 0.42 & \underline{2.34} & \textbf{1.00} & 19.5 \\
3 & \checkmark & \xmark      & \checkmark  & \xmark  & 1.24 & 0.39 & \textbf{2.11} & \textbf{1.00} & 19.1 \\
4 & \checkmark & \checkmark  & \xmark    & \xmark    & 1.63 & \underline{0.23} & 4.16 & \textbf{1.00} & 26.5 \\
5 & \checkmark & \checkmark  & \xmark    & \checkmark    & \underline{0.78} & 0.26 & 3.61 & \textbf{1.00} & 28.5 \\
6 & \checkmark & \checkmark  & \checkmark  & \xmark  & 1.10 & \textbf{0.21} & 3.53 & \textbf{1.00} & \underline{30.8} \\
7 & \checkmark & \checkmark  & \checkmark  & \checkmark  & \textbf{0.52} & 0.27 & 3.78 & \textbf{1.00} & \textbf{31.1} \\
\bottomrule
\end{tabular*}
\end{table}

Point-based contact yields weak stability due to incomplete contact details and hand-crafted rules. Switching to VolCo ($1\rightarrow 2$) obviously improves stability. Adding prior guidance ($2\rightarrow 4; 3\rightarrow 5$) greatly decreases penetration depth, and increases IV and CA, while adding initial guess ($4\rightarrow 5; 6\rightarrow 7$) further improves stability and CA, highlighting the dual role of the auxiliary branch: it regularizes generated VolCo during training to allow being fitted by MANO with less penetrations, and provides better initialization during testing to fit larger contact areas. Interestingly, the stability loss reduce displacement obviously only when there is prior guidance ($4\rightarrow 6; 5\rightarrow 7$), suggesting geometric plausibility is a prerequisite for physical stability.

\subsection{Runtime Analysis}

In this section, we analyze the trade-off between efficiency and performance against prior methods. Table.~\ref{tab:runtime} reports a detailed runtime breakdown, with all experiments run on a single RTX 4090 GPU. Our method achieves the best simulation displacement and penetration depth among all compared methods. VAE-based methods need only a single forward pass at inference, while diffusion-based methods require iterative denoising, so the latter are slower. Among diffusion-based methods, ours has the second-fastest inference time. Its optimization time is also the second fastest overall, behind only the point-contact baseline~\cite{liuTaskOrientedHumanGrasp2025}. Additionally, reducing the resolution from $128\times 8^3$ to $64\times 4^3$ has little effect on efficiency, suggesting the preference for the normal resolution.

However, the point-contact baseline shows obvious drawback in batched inference. Although our method has more parameters (74.04M vs.\ 24.41M) and higher peak memory (358.1M vs.\ 191.4M), it achieves substantially lower TFLOPs (2.01 vs.\ 13.2). The major source of computation is the UNet attention blocks which scales as $O(N)$ in token count, not the volumetric representation due VAE compression. Thus our inference time barely changes as the number of samples increases from 1 to 10 (5.28\unit{\second}$\rightarrow$5.47\unit{\second}), but the time of baseline increases substantially (2.90\unit{\second}$\rightarrow$19.57\unit{\second}).

\begin{table}[t]
\centering
\caption{Per sample runtime breakdown and performance, tested on GRAB dataset. We provide a spars version (VolCoDiff-S) to study the influence of volume resolutions. $^\dagger$GraspTTA is not trained on GRAB, so the SD and PD results are not strictly comparable.}\label{tab:runtime}
\begin{tabular}{l|ccccc|cc}
\toprule
& \thead{Generator \\ Framework} & \thead{Inference \\ time (s)} & \thead{Optim. \\ Iterations} & \thead{Optim.\\ time (s)} & \thead{Total \\ time (s)} & \thead{SD (cm)} & \thead{PD (cm)} \\
\midrule
GrabNet \cite{taheriGRABDatasetWholeBody2020}             & VAE                    & \textbf{0.14}               & -                                             &  -                             & \textbf{0.14}           & 1.02             & 0.40             \\
GraspTTA$^\dagger$  \cite{jiangHandObjectContactConsistency2021}                & VAE                    & 0.17                        & 1000                                          & 6.83                  & 7.00                    & 3.35                & 0.64                \\
ContactGen  \cite{liuContactGenGenerativeContact2023}        & VAE                    & 0.34                        & 1200                                         & 43.6                          & 44.0                   & 2.32             & 0.52             \\
FastGrasp   \cite{wuFastGraspEfficientGrasp2024}              & Diffusion              & 8.60                        & -                                             & -                              & 8.60                    & 2.55             & 0.53             \\
FAGrasp    \cite{chenForceAware3DContact2025}               & VAE                    & 0.415                       & 1200                                         & 39.1                          & 39.5                   & 0.61             & 0.73             \\
Point-Contact Diff \cite{liuTaskOrientedHumanGrasp2025} & Diffusion              & 2.90                        & 1000                                 & \textbf{4.69}                           & 7.59                    & 1.72             & 0.42             \\
\midrule
VolCoDiff-S ($64\times 4^3$)            & Diffusion              & 5.48                        & 1000                                          & 4.89                           & 10.4                   & 0.67             & 0.32             \\
VolCoDiff ($128\times 8^3$)              & Diffusion              & 5.28                        & 1000                                          & 5.39                           & 10.7                   & \textbf{0.52}    & \textbf{0.27}   \\
\bottomrule
\end{tabular}
\end{table}



\section{Conclusion}

We propose VolCo, a novel volumetric contact representation that faithfully preserves fine-grained details in hand-object contact. Building on VolCo, we present VolCoDiff, a volume-wise latent diffusion model, in which the local details are handled by VolumeVAE and the global composition of hand configurations is mitigated by diffusion. The experiments demonstrate the highest precision in reconstruction from contact and the state-of-the-art geometric plausibility and stability in grasp generation, emphasizing the value of introducing hierarchical contact volumes.

\textbf{Limitations}. Our method is currently limited in controllability and scale adaptivity. It prefers only tighter grasps, and the fixed number and scale of volumes limit the coverage of object scales. Future work will explore ways to control VolumeVAE latents and apply adaptive volume numbers and scales.

\textbf{Broader Impact}. Our method can help with improving AR/VR experiences and dexterous robotic deployment, but risks fostering the misuse of robots in harmful contexts.

\section{Acknowledgements}

This research was funded by the China Scholarship Council - University of Birmingham PhD Scholarship programme (No. 202406230091) and the MSIT(Ministry of Science and ICT), Korea, under the ITRC(Information Technology Research Center) support program(IITP-2025-RS-2020-II201789) supervised by the IITP(Institute for Information \& Communications Technology Planning \& Evaluation).

\printbibliography

@String(CVPR  = {IEEE Conf. Comput. Vis. Pattern Recog.})

@String(ICCV  = {Int. Conf. Comput. Vis.})

@String(ECCV  = {Eur. Conf. Comput. Vis.})

@String(ICLR  = {Int. Conf. Learn. Represent.})

@String(AAAI  = {AAAI})

@String(CVPR  = {CVPR})

@String(ICCV  = {ICCV})

@String(ECCV  = {ECCV})

@String(ICLR  = {ICLR})

@inproceedings{liuTaskOrientedHumanGrasp2025,
  title = {Task-{{Oriented Human Grasp Synthesis}} via {{Context-}} and {{Task-Aware Diffusers}}},
  booktitle = {Proceedings of the {{IEEE}}/{{CVF International Conference}} on {{Computer Vision}}},
  author = {Liu, An-Lun and Chao, Yu-Wei and Chen, Yi-Ting},
  year = 2025,
  pages = {10375--10385}
}

@inproceedings{liLatentHOIGeneralizableHand2025,
  title = {{{LatentHOI}}: {{On}} the {{Generalizable Hand Object Motion Generation}} with {{Latent Hand Diffusion}}.},
  shorttitle = {{{LatentHOI}}},
  booktitle = {Proceedings of the {{Computer Vision}} and {{Pattern Recognition Conference}}},
  author = {Li, Muchen and Christen, Sammy and Wan, Chengde and Cai, Yujun and Liao, Renjie and Sigal, Leonid and Ma, Shugao},
  year = 2025,
  pages = {17416--17425},
  doi = {10.1109/CVPR52734.2025.01623}
}

@article{zhangManiDextHandObjectManipulation2025,
  title = {{{ManiDext}}: {{Hand-Object Manipulation Synthesis}} via {{Continuous Correspondence Embeddings}} and {{Residual-Guided Diffusion}}},
  shorttitle = {{{ManiDext}}},
  author = {Zhang, Jiajun and Zhang, Yuxiang and An, Liang and Li, Mengcheng and Zhang, Hongwen and Hu, Zonghai and Liu, Yebin},
  year = 2025,
  journal = {IEEE Transactions on Pattern Analysis and Machine Intelligence},
  pages = {1--15},
  issn = {1939-3539},
  doi = {10.1109/TPAMI.2025.3588302}
}

@inproceedings{yangCPFLearningContact2021,
  title = {{{CPF}}: {{Learning}} a {{Contact Potential Field To Model}} the {{Hand-Object Interaction}}},
  shorttitle = {{{CPF}}},
  booktitle = {Proceedings of the {{IEEE}}/{{CVF International Conference}} on {{Computer Vision}}},
  author = {Yang, Lixin and Zhan, Xinyu and Li, Kailin and Xu, Wenqiang and Li, Jiefeng and Lu, Cewu},
  year = 2021,
  pages = {11097--11106}
}

@inproceedings{yeGHOPGenerativeHandObject2024,
  title = {G-{{HOP}}: {{Generative Hand-Object Prior}} for {{Interaction Reconstruction}} and {{Grasp Synthesis}}},
  shorttitle = {G-{{HOP}}},
  booktitle = {Proceedings of the {{IEEE}}/{{CVF Conference}} on {{Computer Vision}} and {{Pattern Recognition}}},
  author = {Ye, Yufei and Gupta, Abhinav and Kitani, Kris and Tulsiani, Shubham},
  year = 2024,
  pages = {1911--1920},
  doi = {10.48550/arXiv.2404.12383}
}

@inproceedings{karunratanakulGraspingFieldLearning2020,
  title = {Grasping {{Field}}: {{Learning Implicit Representations}} for {{Human Grasps}}},
  shorttitle = {Grasping {{Field}}},
  booktitle = {2020 {{International Conference}} on {{3D Vision}} ({{3DV}})},
  author = {Karunratanakul, Korrawe and Yang, Jinlong and Zhang, Yan and Black, Michael J. and Muandet, Krikamol and Tang, Siyu},
  year = 2020,
  month = nov,
  pages = {333--344},
  issn = {2475-7888},
  doi = {10.1109/3DV50981.2020.00043}
}

@inproceedings{liuPointVoxelCNNEfficient2019,
  title = {Point-{{Voxel CNN}} for {{Efficient 3D Deep Learning}}},
  booktitle = {Advances in {{Neural Information Processing Systems}}},
  author = {Liu, Zhijian and Tang, Haotian and Lin, Yujun and Han, Song},
  year = 2019,
  volume = {32},
  publisher = {Curran Associates, Inc.}
}

@inproceedings{yarivMosaicSDF3DGenerative2024a,
  title = {Mosaic-{{SDF}} for {{3D Generative Models}}},
  booktitle = {Proceedings of the {{IEEE}}/{{CVF Conference}} on {{Computer Vision}} and {{Pattern Recognition}}},
  author = {Yariv, Lior and Puny, Omri and Gafni, Oran and Lipman, Yaron},
  year = 2024,
  pages = {4630--4639}
}

@inproceedings{moralesVersatileDifferentiableHandObject2024,
  title = {A {{Versatile}} and {{Differentiable Hand-Object Interaction Representation}}},
  booktitle = {{{IEEE}}/{{CVF Winter Conference}} on {{Applications}} of {{Computer Vision}} ({{WACV}})},
  author = {Morales, Th{\'e}o and Taheri, Omid and Lacey, Gerard},
  year = 2024,
  month = nov,
  eprint = {2409.16855},
  primaryclass = {cs},
  publisher = {IEEE/CVF},
  doi = {10.1109/WACV61041.2025.00013},
  archiveprefix = {arXiv}
}

@inproceedings{gradyContactOptOptimizingContact2021,
  title = {{{ContactOpt}}: {{Optimizing Contact}} to {{Improve Grasps}}},
  shorttitle = {{{ContactOpt}}},
  booktitle = {2021 {{IEEE}}/{{CVF Conference}} on {{Computer Vision}} and {{Pattern Recognition}} ({{CVPR}})},
  author = {Grady, Patrick and Tang, Chengcheng and Twigg, Christopher D. and Vo, Minh and Brahmbhatt, Samarth and Kemp, Charles C.},
  year = 2021,
  month = jun,
  pages = {1471--1481},
  publisher = {IEEE},
  address = {Nashville, TN, USA},
  doi = {10.1109/CVPR46437.2021.00152},
  copyright = {https://ieeexplore.ieee.org/Xplorehelp/downloads/license-information/IEEE.html},
  isbn = {978-1-6654-4509-2}
}

@inproceedings{jiangHandObjectContactConsistency2021,
  title = {Hand-{{Object Contact Consistency Reasoning}} for {{Human Grasps Generation}}},
  booktitle = {Proceedings of the {{IEEE}}/{{CVF International Conference}} on {{Computer Vision}}},
  author = {Jiang, Hanwen and Liu, Shaowei and Wang, Jiashun and Wang, Xiaolong},
  year = 2021,
  pages = {11107--11116},
  doi = {10.48550/arXiv.2104.03304}
}

@incollection{tseS$^2$ContactGraphBasedNetwork2022,
  title = {S\textasciicircum 2{{Contact}}: {{Graph-Based Network}} for {{3D Hand-Object Contact Estimation}} with {{Semi-supervised Learning}}},
  shorttitle = {S\$\$\textasciicircum 2\$\${{Contact}}},
  booktitle = {Computer {{Vision}} -- {{ECCV}} 2022},
  author = {Tse, Tze Ho Elden and Zhang, Zhongqun and Kim, Kwang In and Leonardis, Ale{\u s} and Zheng, Feng and Chang, Hyung Jin},
  editor = {Avidan, Shai and Brostow, Gabriel and Ciss{\'e}, Moustapha and Farinella, Giovanni Maria and Hassner, Tal},
  year = 2022,
  volume = {13661},
  pages = {568--584},
  publisher = {Springer Nature Switzerland},
  address = {Cham},
  doi = {10.1007/978-3-031-19769-7_33},
  isbn = {978-3-031-19768-0}
}

@inproceedings{liuContactGenGenerativeContact2023,
  title = {{{ContactGen}}: {{Generative Contact Modeling}} for {{Grasp Generation}}},
  shorttitle = {{{ContactGen}}},
  booktitle = {2023 {{IEEE}}/{{CVF International Conference}} on {{Computer Vision}} ({{ICCV}})},
  author = {Liu, Shaowei and Zhou, Yang and Yang, Jimei and Gupta, Saurabh and Wang, Shenlong},
  year = 2023,
  month = oct,
  pages = {20552--20563},
  publisher = {IEEE},
  address = {Paris, France},
  doi = {10.1109/ICCV51070.2023.01884},
  copyright = {https://doi.org/10.15223/policy-029},
  isbn = {979-8-3503-0718-4}
}

@misc{neverovaContinuousSurfaceEmbeddings2020,
  title = {Continuous {{Surface Embeddings}}},
  author = {Neverova, Natalia and Novotny, David and Khalidov, Vasil and Szafraniec, Marc and Labatut, Patrick and Vedaldi, Andrea},
  year = 2020,
  month = nov,
  number = {arXiv:2011.12438},
  eprint = {2011.12438},
  primaryclass = {cs},
  publisher = {arXiv},
  doi = {10.48550/arXiv.2011.12438},
  archiveprefix = {arXiv}
}

@inproceedings{chenForceAware3DContact2025,
  title = {Force-{{Aware 3D Contact Modeling}} for {{Stable Grasp Generation}}},
  booktitle = {Proceedings of the {{AAAI Conference}} on {{Artificial Intelligence}}},
  author = {Chen, Zhuo and Zhang, Zhongqun and Cheng, Yihua and Leonardis, Ales and Chang, Hyung Jin},
  year = 2025,
  month = nov,
  eprint = {2511.13247},
  primaryclass = {cs},
  doi = {10.48550/arXiv.2511.13247},
}

@article{romeroEmbodiedHandsModeling2017,
  title = {Embodied {{Hands}}: {{Modeling}} and {{Capturing Hands}} and {{Bodies Together}}},
  shorttitle = {Embodied {{Hands}}},
  author = {Romero, Javier and Tzionas, Dimitrios and Black, Michael J.},
  year = 2017,
  month = dec,
  journal = {ACM Transactions on Graphics},
  volume = {36},
  number = {6},
  eprint = {2201.02610},
  primaryclass = {cs},
  pages = {1--17},
  issn = {0730-0301, 1557-7368},
  doi = {10.1145/3130800.3130883},
  archiveprefix = {arXiv}
}

@article{zuoGraspDiffGraspingGeneration2024,
  title = {{{GraspDiff}}: {{Grasping Generation}} for {{Hand-Object Interaction With Multimodal Guided Diffusion}}},
  shorttitle = {{{GraspDiff}}},
  author = {Zuo, Binghui and Zhao, Zimeng and Sun, Wenqian and Yuan, Xiaohan and Yu, Zhipeng and Wang, Yangang},
  year = 2024,
  journal = {IEEE Transactions on Visualization and Computer Graphics},
  pages = {1--13},
  issn = {1941-0506},
  doi = {10.1109/TVCG.2024.3466190}
}

@inproceedings{kingmaAutoEncodingVariationalBayes2014,
  title = {Auto-{{Encoding Variational Bayes}}},
  booktitle = {Proceedings of the 2nd {{International Conference}} on {{Learning Representations}} ({{ICLR}})},
  author = {Kingma, Diederik P. and Welling, Max},
  editor = {Bengio, Yoshua and LeCun, Yann},
  year = 2014,
  doi = {10.48550/arXiv.1312.6114}
}

@inproceedings{hoDenoisingDiffusionProbabilistic2020,
  title = {Denoising {{Diffusion Probabilistic Models}}},
  booktitle = {Advances in {{Neural Information Processing Systems}}},
  author = {Ho, Jonathan and Jain, Ajay and Abbeel, Pieter},
  year = 2020,
  volume = {33},
  pages = {6840--6851},
  publisher = {Curran Associates, Inc.},
  doi = {10.48550/arXiv.2006.11239}
}

@inproceedings{karunratanakulSkeletonDrivenNeuralOccupancy2021,
  title = {A {{Skeleton-Driven Neural Occupancy Representation}} for {{Articulated Hands}}},
  booktitle = {Proceedings of the {{International Conference}} on {{3D Vision}}},
  author = {Karunratanakul, Korrawe and Spurr, Adrian and Fan, Zicong and Hilliges, Otmar and Tang, Siyu},
  year = 2021,
  month = sep,
  eprint = {2109.11399},
  publisher = {IEEE},
  doi = {10.48550/arXiv.2109.11399},
  archiveprefix = {arXiv}
}

@inproceedings{bowman2016generating,
  title={Generating sentences from a continuous space},
  author={Bowman, Samuel R and Vilnis, Luke and Vinyals, Oriol and Dai, Andrew M and Jozefowicz, Rafal and Bengio, Samy},
  booktitle={20th SIGNLL Conference on Computational Natural Language Learning, CoNLL 2016},
  pages={10--21},
  year={2016},
  organization={Association for Computational Linguistics (ACL)}
}

@inproceedings{shimadaMACSMassConditioned2024,
  title = {{{MACS}}: {{Mass Conditioned 3D Hand}} and {{Object Motion Synthesis}}},
  shorttitle = {{{MACS}}},
  booktitle = {2024 {{International Conference}} on {{3D Vision}} ({{3DV}})},
  author = {Shimada, Soshi and Mueller, Franziska and Bednarik, Jan and Doosti, Bardia and Bickel, Bernd and Tang, Danhang and Golyanik, Vladislav and Taylor, Jonathan and Theobalt, Christian and Beeler, Thabo},
  year = 2024,
  month = mar,
  pages = {1082--1091},
  issn = {2475-7888},
  doi = {10.1109/3DV62453.2024.00082}
}

@inproceedings{taheriGRABDatasetWholeBody2020,
  title = {{{GRAB}}: {{A Dataset}} of {{Whole-Body Human Grasping}} of {{Objects}}},
  shorttitle = {{{GRAB}}},
  booktitle = {Computer {{Vision}} -- {{ECCV}} 2020},
  author = {Taheri, Omid and Ghorbani, Nima and Black, Michael J. and Tzionas, Dimitrios},
  editor = {Vedaldi, Andrea and Bischof, Horst and Brox, Thomas and Frahm, Jan-Michael},
  year = 2020,
  pages = {581--600},
  publisher = {Springer International Publishing},
  address = {Cham},
  doi = {10.1007/978-3-030-58548-8_34},
  isbn = {978-3-030-58548-8}
}

@inproceedings{zhengCAMSCAnonicalizedManipulation2023,
  title = {{{CAMS}}: {{CAnonicalized Manipulation Spaces}} for {{Category-Level Functional Hand-Object Manipulation Synthesis}}},
  shorttitle = {{{CAMS}}},
  booktitle = {Proceedings of the {{IEEE}}/{{CVF Conference}} on {{Computer Vision}} and {{Pattern Recognition}}},
  author = {Zheng, Juntian and Zheng, Qingyuan and Fang, Lixing and Liu, Yun and Yi, Li},
  year = 2023,
  pages = {585--594},
  doi = {10.48550/arXiv.2303.15469}
}

@inproceedings{rombachHighResolutionImageSynthesis2022,
  title = {High-{{Resolution Image Synthesis With Latent Diffusion Models}}},
  booktitle = {Proceedings of the {{IEEE}}/{{CVF Conference}} on {{Computer Vision}} and {{Pattern Recognition}}},
  author = {Rombach, Robin and Blattmann, Andreas and Lorenz, Dominik and Esser, Patrick and Ommer, Bj{\"o}rn},
  year = 2022,
  pages = {10684--10695}
}

@inproceedings{chaText2HOITextguided3D2024,
  title = {{{Text2HOI}}: {{Text-guided 3D Motion Generation}} for {{Hand-Object Interaction}}},
  shorttitle = {{{Text2HOI}}},
  booktitle = {Proceedings of the {{IEEE}}/{{CVF Conference}} on {{Computer Vision}} and {{Pattern Recognition}}},
  author = {Cha, Junuk and Kim, Jihyeon and Yoon, Jae Shin and Baek, Seungryul},
  year = 2024,
  pages = {1577--1585},
  doi = {10.48550/arXiv.2404.00562}
}

@inproceedings{chenGSDFGeometryDrivenSigned2023,
  title = {{{gSDF}}: {{Geometry-Driven Signed Distance Functions}} for {{3D Hand-Object Reconstruction}}},
  shorttitle = {{{gSDF}}},
  booktitle = {Proceedings of the {{IEEE}}/{{CVF Conference}} on {{Computer Vision}} and {{Pattern Recognition}}},
  author = {Chen, Zerui and Chen, Shizhe and Schmid, Cordelia and Laptev, Ivan},
  year = 2023,
  pages = {12890--12900}
}

@inproceedings{zhangNL2ContactNaturalLanguage2024,
  title = {{{NL2Contact}}: {{Natural Language Guided 3D Hand-Object Contact Modeling}} with {{Diffusion Model}}},
  shorttitle = {{{NL2Contact}}},
  booktitle = {Proceedings of the 2024 {{European Conference}} on {{Computer Vision}}},
  author = {Zhang, Zhongqun and Wang, Hengfei and Yu, Ziwei and Cheng, Yihua and Yao, Angela and Chang, Hyung Jin},
  year = 2024,
  month = jul,
  eprint = {2407.12727},
  primaryclass = {cs},
  publisher = {ECVA},
  doi = {10.48550/arXiv.2407.12727},
  archiveprefix = {arXiv}
}

@incollection{zhouTOCHSpatioTemporalObjecttoHand2022,
  title = {{{TOCH}}: {{Spatio-Temporal Object-to-Hand Correspondence}} for {{Motion Refinement}}},
  shorttitle = {{{TOCH}}},
  booktitle = {Computer {{Vision}} -- {{ECCV}} 2022},
  author = {Zhou, Keyang and Bhatnagar, Bharat Lal and Lenssen, Jan Eric and {Pons-Moll}, Gerard},
  editor = {Avidan, Shai and Brostow, Gabriel and Ciss{\'e}, Moustapha and Farinella, Giovanni Maria and Hassner, Tal},
  year = 2022,
  volume = {13663},
  pages = {1--19},
  publisher = {Springer Nature Switzerland},
  address = {Cham},
  doi = {10.1007/978-3-031-20062-5_1},
  isbn = {978-3-031-20061-8}
}

@inproceedings{cicek3DUNetLearning2016,
  title = {{{3D U-Net}}: {{Learning Dense Volumetric Segmentation}} from {{Sparse Annotation}}},
  shorttitle = {{{3D U-Net}}},
  booktitle = {Medical {{Image Computing}} and {{Computer-Assisted Intervention}} -- {{MICCAI}} 2016: 19th {{International Conference}}, {{Athens}}, {{Greece}}, {{October}} 17-21, 2016, {{Proceedings}}, {{Part II}}},
  author = {{\c C}i{\c c}ek, {\"O}zg{\"u}n and Abdulkadir, Ahmed and Lienkamp, Soeren S. and Brox, Thomas and Ronneberger, Olaf},
  year = 2016,
  month = oct,
  pages = {424--432},
  publisher = {Springer-Verlag},
  address = {Berlin, Heidelberg},
  doi = {10.1007/978-3-319-46723-8_49},
  isbn = {978-3-319-46722-1}
}

@inproceedings{brockGenerativeDiscriminativeVoxel2016,
  title = {Generative and {{Discriminative Voxel Modeling}} with {{Convolutional Neural Networks}}},
  booktitle = {Advances in {{Neural Information Processing Systems}}},
  author = {Brock, Andrew and Lim, Theodore and Ritchie, J. M. and Weston, Nick},
  year = 2016,
  month = aug,
  eprint = {1608.04236},
  primaryclass = {cs},
  doi = {10.48550/arXiv.1608.04236},
  archiveprefix = {arXiv}
}

@inproceedings{hampaliHOnnotateMethod3D2020,
  title = {{{HOnnotate}}: {{A Method}} for {{3D Annotation}} of {{Hand}} and {{Object Poses}}},
  shorttitle = {{{HOnnotate}}},
  booktitle = {Proceedings of the {{IEEE}}/{{CVF Conference}} on {{Computer Vision}} and {{Pattern Recognition}}},
  author = {Hampali, Shreyas and Rad, Mahdi and Oberweger, Markus and Lepetit, Vincent},
  year = 2020,
  pages = {3196--3206},
  doi = {10.48550/arXiv.1907.01481}
}

@inproceedings{huang2023diffusion,
  title={Diffusion-based Generation, Optimization, and Planning in 3D Scenes},
  author={Huang, Siyuan and Wang, Zan and Li, Puhao and Jia, Baoxiong and Liu, Tengyu and Zhu, Yixin and Liang, Wei and Zhu, Song-Chun},
  booktitle={Proceedings of the IEEE/CVF Conference on Computer Vision and Pattern Recognition (CVPR)},
  year={2023}
}

@inproceedings{qi2017pointnet,
  title={Pointnet: Deep learning on point sets for 3d classification and segmentation},
  author={Qi, Charles R and Su, Hao and Mo, Kaichun and Guibas, Leonidas J},
  booktitle={Proceedings of the IEEE conference on computer vision and pattern recognition},
  pages={652--660},
  year={2017}
}

@inproceedings{tevetHumanMotionDiffusion2022,
  title = {Human {{Motion Diffusion Model}}},
  booktitle = {The {{Eleventh International Conference}} on {{Learning Representations}}},
  author = {Tevet, Guy and Raab, Sigal and Gordon, Brian and Shafir, Yonatan and {Cohen-Or}, Daniel and Bermano, Amit H.},
  year = 2022,
  month = oct,
  eprint = {2209.14916},
  primaryclass = {cs},
  publisher = {ICLR},
  doi = {10.48550/arXiv.2209.14916},
  archiveprefix = {arXiv}
}

@inproceedings{vaswaniAttentionAllYou2017,
  title = {Attention Is {{All}} You {{Need}}},
  booktitle = {Advances in {{Neural Information Processing Systems}}},
  author = {Vaswani, Ashish and Shazeer, Noam and Parmar, Niki and Uszkoreit, Jakob and Jones, Llion and Gomez, Aidan N and ukasz Kaiser, {\L} and Polosukhin, Illia},
  year = 2017,
  volume = {30},
  publisher = {Curran Associates, Inc.}
}

@inproceedings{zimmermannFreiHANDDatasetMarkerless2019,
  title = {{{FreiHAND}}: {{A Dataset}} for {{Markerless Capture}} of {{Hand Pose}} and {{Shape From Single RGB Images}}},
  shorttitle = {{{FreiHAND}}},
  booktitle = {Proceedings of the {{IEEE}}/{{CVF International Conference}} on {{Computer Vision}}},
  author = {Zimmermann, Christian and Ceylan, Duygu and Yang, Jimei and Russell, Bryan and Argus, Max and Brox, Thomas},
  year = 2019,
  pages = {813--822}
}

@article{FAWCETT2006861,
  title = {An Introduction to {{ROC}} Analysis},
  author = {Fawcett, Tom},
  year = 2006,
  journal = {Pattern Recognition Letters},
  volume = {27},
  number = {8},
  pages = {861--874},
  issn = {0167-8655},
  doi = {10.1016/j.patrec.2005.10.010}
}

@inproceedings{xuDexterousGraspTransformer2024,
  title = {Dexterous {{Grasp Transformer}}},
  booktitle = {Proceedings of the {{IEEE}}/{{CVF Conference}} on {{Computer Vision}} and {{Pattern Recognition}}},
  author = {Xu, Guo-Hao and Wei, Yi-Lin and Zheng, Dian and Wu, Xiao-Ming and Zheng, Wei-Shi},
  year = 2024,
  pages = {17933--17942},
  doi = {10.48550/arXiv.2404.18135}
}

@article{hunt1975coefficient,
  title={Coefficient of restitution interpreted as damping in vibroimpact},
  author={Hunt, K. H. and Crossley, F. R. E.},
  journal={Journal of Applied Mechanics},
  volume={42},
  number={2},
  pages={440--445},
  year={1975},
  publisher={American Society of Mechanical Engineers}
}

@inproceedings{wuFastGraspEfficientGrasp2024,
  title = {{{FastGrasp}}: {{Efficient Grasp Synthesis}} with {{Diffusion}}},
  shorttitle = {{{FastGrasp}}},
  booktitle = {International {{Conference}} on {{3D Vision}} ({{3DV}})},
  author = {Wu, Xiaofei and Liu, Tao and Li, Caoji and Ma, Yuexin and Shi, Yujiao and He, Xuming},
  year = 2024,
  month = nov,
  eprint = {2411.14786},
  primaryclass = {cs},
  doi = {10.48550/arXiv.2411.14786},
  archiveprefix = {arXiv}
}

@MISC{coumans2021,
author =   {Erwin Coumans and Yunfei Bai},
title =    {PyBullet, a Python module for physics simulation for games, robotics and machine learning},
howpublished = {\url{http://pybullet.org}},
year = {2016--2021}
}

@INPROCEEDINGS{grady2024Pressurevisionpp,
  author={Grady, Patrick and Collins, Jeremy A. and Tang, Chengcheng and Twigg, Christopher D. and Aneja, Kunal and Hays, James and Kemp, Charles C.},
  booktitle={2024 IEEE/CVF Winter Conference on Applications of Computer Vision (WACV)}, 
  title={PressureVision++: Estimating Fingertip Pressure from Diverse RGB Images}, 
  year={2024},
  volume={},
  number={},
  pages={8683-8693},
doi={10.1109/WACV57701.2024.00850}}

@article{Li2024ShapeGraspZT,
title={ShapeGrasp: Zero-Shot Task-Oriented Grasping with Large Language Models through Geometric Decomposition},
author={Samuel Li and Sarthak Bhagat and Joseph Campbell and Yaqi Xie and Woojun Kim and Katia P. Sycara and Simon Stepputtis},
journal={2024 IEEE/RSJ International Conference on Intelligent Robots and Systems (IROS)},
year={2024},
pages={10527-10534},
}

@article{guo2021human,
  title={Human-machine interaction sensing technology based on hand gesture recognition: A review},
  author={Guo, Lin and Lu, Zongxing and Yao, Ligang},
  journal={IEEE Transactions on Human-Machine Systems},
  pages={300--309},
  year={2021},
  publisher={IEEE}
}

@misc{yang2025egovlalearningvisionlanguageactionmodels,
    title={EgoVLA: Learning Vision-Language-Action Models from Egocentric Human Videos}, 
    author={Ruihan Yang and Qinxi Yu and Yecheng Wu and Rui Yan and Borui Li and An-Chieh Cheng and Xueyan Zou and Yunhao Fang and Hongxu Yin and Sifei Liu and Song Han and Yao Lu and Xiaolong Wang},
    year={2025},
    eprint={2507.12440},
    archivePrefix={arXiv},
    primaryClass={cs.RO},
    url={https://arxiv.org/abs/2507.12440}, 
  }

@inproceedings{chaoDexYCBBenchmarkCapturing2021,
  title = {{{DexYCB}}: {{A Benchmark}} for {{Capturing Hand Grasping}} of {{Objects}}},
  shorttitle = {{{DexYCB}}},
  booktitle = {Proceedings of the {{IEEE}}/{{CVF Conference}} on {{Computer Vision}} and {{Pattern Recognition}}},
  author = {Chao, Yu-Wei and Yang, Wei and Xiang, Yu and Molchanov, Pavlo and Handa, Ankur and Tremblay, Jonathan and Narang, Yashraj S. and Van Wyk, Karl and Iqbal, Umar and Birchfield, Stan and Kautz, Jan and Fox, Dieter},
  year = 2021,
  pages = {9044--9053},
  doi = {10.48550/arXiv.2104.04631}
}

@article{loshchilov2017decoupled,
  title={Decoupled weight decay regularization},
  author={Loshchilov, Ilya and Hutter, Frank},
  journal={arXiv preprint arXiv:1711.05101},
  year={2017}
}

@article{qi2017pointnetplusplus,
  title={PointNet++: Deep Hierarchical Feature Learning on Point Sets in a Metric Space},
  author={Qi, Charles R and Yi, Li and Su, Hao and Guibas, Leonidas J},
  journal={arXiv preprint arXiv:1706.02413},
  year={2017}
}

\appendix

\section{Stability Loss With Slack Variables}
\label{app:stability_loss}

In this section, we briefly restate the stability loss in the literature \cite{chenForceAware3DContact2025}, and how it gets improved by slack variables.

Suppose a grasp with an object of mass $m$, inertia matrix $I$, and surface friction ratio $\mu$ in a scene with gravitational acceleration $\mathbf{g}$. For a contact point $\mathbf{x}$, the normal force applied is denoted by $\mathbf{F}_{\mathbf{x}} = F_{\mathbf{x}}\mathbf{n}_{\mathbf{x}}\in\mathbb{R}^3$, where $\mathbf{n}_{\mathbf{x}}$ is the normal direction. With the object SDF, the direction is formally defined as $\mathbf{n}_{\mathbf{x}}=-\dot{\text{SDF}}_o(\mathbf{x})$. Suppose $\mathbf{b}_{\mathbf{x}}, \mathbf{t}_{\mathbf{x}}, \mathbf{n}_{\mathbf{x}}$ form a regular basis and $\bar{\mathbf{b}}_{\mathbf{x}} = (\mathbf{x} - \mathbf{p}_{CoM})\times \mathbf{b}_\mathbf{x}$ \etc, then the rotational and angular accelerations are calculated by

\begin{eqnarray}
    \mathbf{a} &=& \mathbf{g} + \frac{1}{m}\sum_{i=1}^N\sum_{\mathbf{x}\in\mathcal{G}_i}F_{\mathbf{x}}\left(\mathbf{n}_{\mathbf{x}} + \mu\delta_\mathbf{x}\mathbf{b}_{\mathbf{x}} + \mu\gamma_\mathbf{x}\mathbf{t}_{\mathbf{x}}\right); \\
    \boldsymbol{\alpha} &=& I^{-1}\sum_{i=1}^N\sum_{\mathbf{x}\in\mathcal{G}_i}F_{\mathbf{x}}\left(\bar{\mathbf{n}}_{\mathbf{x}} + \mu\delta_\mathbf{x}\bar{\mathbf{b}}_{\mathbf{x}} + \mu\gamma_\mathbf{x}\bar{\mathbf{t}}_{\mathbf{x}}\right),
\end{eqnarray}
where $\delta_{\mathbf{x}}, \gamma_{\mathbf{x}} \in [-1, 1]$ are variables to determine friction.

The stability loss is ideally 

\begin{equation}\label{eq:lstable0}
    E_{stability} = \|\mathbf{a}\|^2 + \|\boldsymbol{\alpha}\|^2.
\end{equation}

Following the paper \cite{chenForceAware3DContact2025}, to formulate Eq.~\eqref{eq:lstable0} to a differentiable loss term, $\delta_{\mathbf{x}}, \gamma_{\mathbf{x}}$ should be eliminated. This is achieved by comparing the possible upper and lower bounds of each dimension of the acceleration. Use the subscript $j=1, 2, 3$ to represent the $j$\textsuperscript{th} element in the vector, then we have

\begin{multline}\label{eq:approx_stable_a}
    \mathbf{g}_j + \frac{1}{m}\sum_{i=1}^N\sum_{\mathbf{x}\in\mathcal{G}_i} F_{\mathbf{x}}\left(\mathbf{n}_{\mathbf{x}j} - \mu|\mathbf{b}_{\mathbf{x}j}| - \mu|\mathbf{t}_{\mathbf{x}j}|\right) \overset{\triangle}{=} L_j \leq \mathbf{a}_j \\
    \leq U_j \overset{\triangle}{=} \mathbf{g}_j + \frac{1}{m}\sum_{i=1}^N\sum_{\mathbf{x}\in\mathcal{G}_i} F_{\mathbf{x}}\left(\mathbf{n}_{\mathbf{x}j} + \mu|\mathbf{b}_{\mathbf{x}j}| + \mu|\mathbf{t}_{\mathbf{x}j}|\right).
    \end{multline}
    
\begin{multline}\label{eq:approx_stable_alpha}
    \sum_{i=1}^N\sum_{\mathbf{x}\in\mathcal{G}_i} F_{\mathbf{x}}\left(I^{-1}_{j\cdot}\bar{\mathbf{n}}_{\mathbf{x}} - \mu|I^{-1}_{j\cdot}\bar{\mathbf{b}}_{\mathbf{x}}| - \mu|I^{-1}_{j\cdot}\bar{\mathbf{t}}_{\mathbf{x}}|\right)\overset{\triangle}{=}\mathcal{L}_j \leq \boldsymbol{\alpha}_j \\
    \leq \mathcal{U}_j \overset{\triangle}{=} \sum_{i=1}^N\sum_{\mathbf{x}\in\mathcal{G}_i} F_{\mathbf{x}}\left(I^{-1}_{j\cdot}\bar{\mathbf{n}}_{\mathbf{x}} + \mu|I^{-1}_{j\cdot}\bar{\mathbf{b}}_{\mathbf{x}}| + \mu|I^{-1}_{j\cdot}\bar{\mathbf{t}}_{\mathbf{x}}|\right).
\end{multline}

So the stability loss is designed to punish only the cases when 0 fall out of the feasible range.

\begin{equation}\label{eq:stab_loss}
    \mathcal{L}_{stability} = \sum_{j=1}^3\left(\max\{L_j, 0\} + \max\{\mathcal{L}_j, 0\} - \min\{U_j, 0\} - \min\{\mathcal{U}_j, 0\}\right)
\end{equation}

This formulation, however, cannot be directly applied in our setting. The key concern is that this formula assumes the forces are precise, yet the spring-damper force model $F_\mathbf{x}=k d$ is an approximation and random errors in penetration depth in motion capture data are unavoidable. As stated in the main paper, our solution is to add a slack variable to loosen the constraint on penetration depth, so any depth within $\pm\Delta d$ error is considered correct. In other words, the upper and lower bounds in Eq.~\eqref{eq:approx_stable_a} and \eqref{eq:approx_stable_alpha} can be expanded by replacing the constant $F_{\mathbf{x}}$ with a function of $\Delta d_{\mathbf{x}}$, \eg,

\begin{eqnarray}
    L_j(\Delta d) &=& \mathbf{g}_j + \frac{1}{m}\sum_{i=1}^N\sum_{\mathbf{x}\in\mathcal{G}_i} k(d_\mathbf{x} + \Delta d_{\mathbf{x}})\left(\mathbf{n}_{\mathbf{x}j} - \mu|\mathbf{b}_{\mathbf{x}j}| - \mu|\mathbf{t}_{\mathbf{x}j}|\right) \\
     &=& L_j + \frac{1}{m}\sum_{i=1}^N\sum_{\mathbf{x}\in\mathcal{G}_i} k\Delta d_{\mathbf{x}}\left(\mathbf{n}_{\mathbf{x}j} - \mu|\mathbf{b}_{\mathbf{x}j}| - \mu|\mathbf{t}_{\mathbf{x}j}|\right) \\
     &\geq& L_j - \frac{1}{m}\sum_{i=1}^N\sum_{\mathbf{x}\in\mathcal{G}_i} k\Delta d\left|\mathbf{n}_{\mathbf{x}j} - \mu|\mathbf{b}_{\mathbf{x}j}| - \mu|\mathbf{t}_{\mathbf{x}j}|\right| \overset{\triangle}{=} L_j^\prime.
\end{eqnarray}
The same works for $U_j^\prime, \mathcal{L}_j^\prime, \mathcal{U}_j^\prime$.

Substitute $L_j, \mathcal{L}_j, U_j, \mathcal{U}_j$ with $L_j^\prime, \mathcal{L}_j^\prime, U_j^\prime, \mathcal{U}_j^\prime$ in Eq.~\eqref{eq:stab_loss}, then we get a new stability loss which can be applied to our case.

We set the limit of the slack variables to $\Delta d = 1\unit{\milli\metre}$.

\section{Model Architectures}
\label{app:model}

This section provide details regarding the architectures of VolumeVAE, VolCoDiff UNet, and hand VAE, in support for reproducing the results. 

\subsection{VolumeVAE}

\begin{figure}[t]
    \centering
    \includegraphics[width=\linewidth]{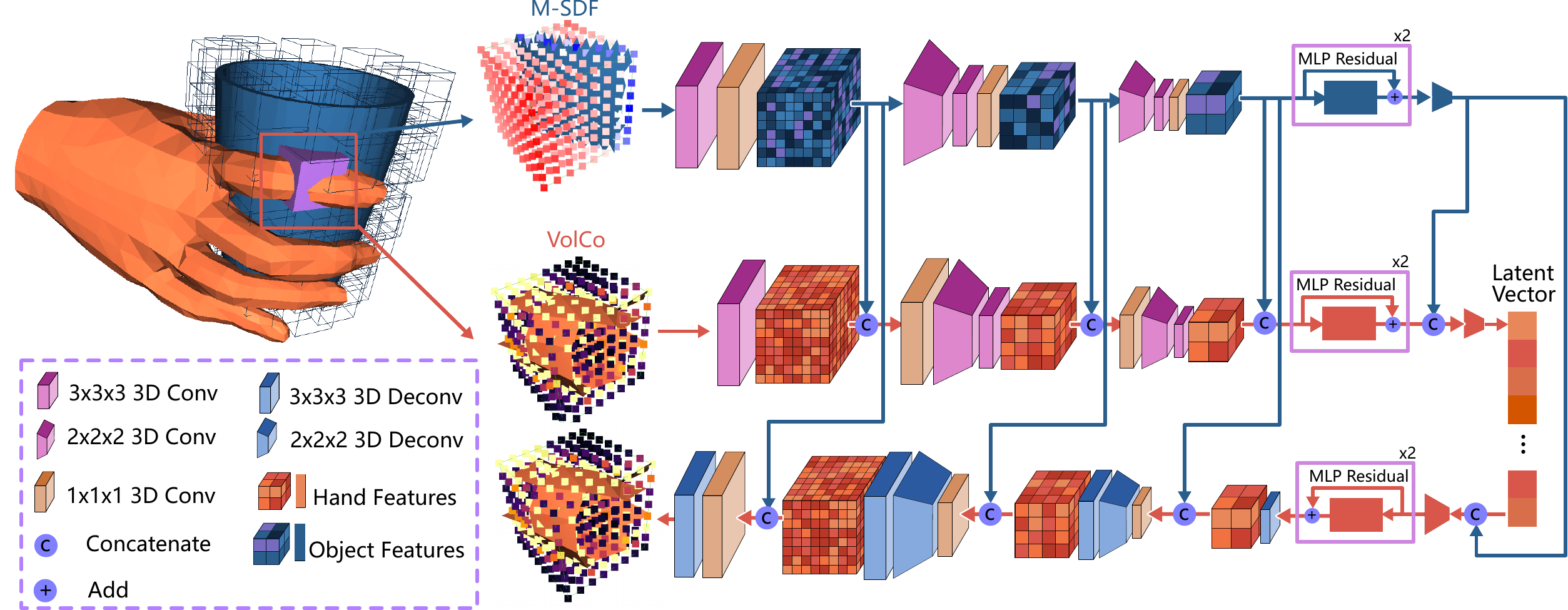}
    \caption{VolumeVAE architecture}
    \label{fig:vvae}
\end{figure}

VolumeVAE adopts a symmetric 3D VAE architecture, illustrated in Fig.~\ref{fig:vvae}. The input VolCo and object SDF volumes are passed through two downsampling stages, flattened, and processed by residual MLP blocks followed by a projection layer to yield a 1D feature vector. Multi-scale object features are injected into the contact features at corresponding scales via channel concatenations. Due to the small input dimensionality ($8^3 \times n_{channel}$), the input is designed to minimize information loss: downsampling layers are implemented by 3D convolutional layers with a kernel size of 2, and all activations are LeakyReLU with a negative slope of 0.2.

\subsection{VolCoDiff}

\begin{figure}[t]
    \centering
    \includegraphics[width=\linewidth]{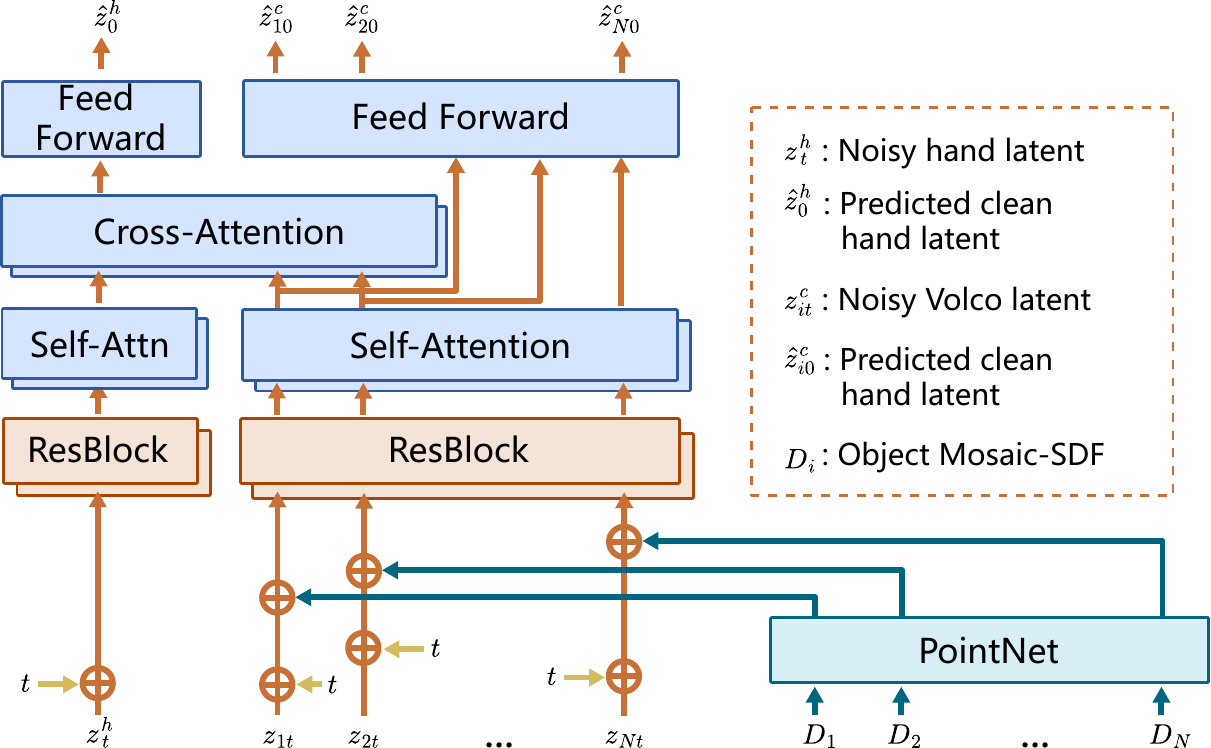}
    \caption{The UNet architecture of VolCoDiff.}
    \label{fig:vcdiff}
\end{figure}

The UNet of VolCoDiff is shown in Fig.~\ref{fig:vcdiff}. The object SDF volumes are flattened to a $8^3=512$-dimensional vector, which are used as feature for each object volume center and processed by a PointNet~\cite{qi2017pointnet} encoder. The hand branch receives the information from the object and the contact by cross-attention, while the contact branch does not receive information from the hand branch. 

\subsection{Hand VAE}

The hand VAE consists of a MLP encoder that receives the flattened hand vertex positions as input and outputs a 32-dimensional latent vector, where the first 16 dimensions represent the mean and the remaining 16 the log-standard-deviation. The latent gets reparameterized to a sampled 16-dimensional latent and decoded by a 2-layer MLP decoder to a 61-dimensional vector $(\mathbf{t} \mid \boldsymbol{\theta} \mid \boldsymbol{\beta})$ representing the MANO parameter of the hand. 

The training of hand VAE is done on GRAB training set. The loss terms include a reconstruction loss $\mathcal{L}_{rec}$ and KL-divergence loss. The reconstruction loss is calculated based on the hand vertices that are predicted from MANO $\{\mathbf{v}_i\}_{i=1}^{778}=\text{MANO}(\mathbf{t} \mid \boldsymbol{\theta} \mid \boldsymbol{\beta})$, \ie

\begin{equation}
\mathcal{L}_{rec} = \frac{1}{778}\sum_{i=1}^{778}MSE(\mathbf{v}_i, \hat{\mathbf{v}}_i)
\end{equation}

With the KL-divergence loss $\mathcal{L}_{KL}=D_{KL}(\mathcal{N}(\boldsymbol{\mu}, \boldsymbol{\sigma}^2) || \mathcal{N}(0, I))$, the loss to train the hand VAE is simply

\begin{equation}
\mathcal{L}_{handvae} = \mathcal{L}_{rec} + \lambda_{KL}\mathcal{L}_{KL}
\end{equation}
where $\lambda_{KL} = 10^{-6}$. 

The model is trained with AdamW optimizer \cite{loshchilov2017decoupled} and $10^{-4}$ learning rate for 100 epochs.

\section{Experimental Details}
\label{app:exp_detail}

This section provides details to reproduce our result in the main paper. Our experiments are done on a linux machine with Intel(R) Xeon(R) w7-3445 CPU, 64GB RAM, and two 4090 GPUs.

\subsection{Data Pre-processing}
\label{app:preprocess}

For training, we adopt the GRAB dataset \cite{taheriGRABDatasetWholeBody2020} with the same pre-processing pipeline as GrabNet, yielding 323,622 frames of closely interacting hand-object pairs. The validation and test sets are obtained by downsampling every 4 and 64 frames, comprising 7,757 and 1,012 frames, respectively. The training set is used directly for training the diffusion model, and the test set is used in the experiment of reconstruction from contact. Further, the training set is augmented by a randomly selected rotation from all 24 axis-aligned ones which is applied to the local volumes. This keeps the rotated volumes still axis-aligned so no interpolations are needed. The resulting Mosaic-SDF and VolCo can be efficiently obtained by reordering the grid points.

However, the training of VolumeVAE requires special attention. The key concern is that one training sample of VolumeVAE should be one single volume with object SDF and contact values. So we construct a local volume dataset based on the sampled points. Each object surface is sampled 1,024 points, so totally 331.4M local volumes are available for training. The dataset exceeds the capacity of a compact VAE model, and a substantial portion of the data carries redundant or repeated information. We therefore apply two-level downsampling during data pre-processing.

To mitigate the similar poses between continuous frames, we downsample the training set every 5 frames, and validation and test set every 10 frames; To reduce the redundancy of nearby sampled points, we also downsample the object points each frame by 40\% (training set) and 20\% (validation and testing). Since in our VolCoDiff, non-contact volumes are set to 0, so fitting to those volumes in training is not necessary. Thus, we only take the sampled volumes with at least one hand vertex inside to construct the dataset. Finally, we arrive at a training set of 7,320,315 volumes, a validation set of 165,651 and a test set of 273,315.

\subsection{Contact-Map Based Method}

The contact-map based method that we utilize in ablation study follows the architecture of \cite{liuTaskOrientedHumanGrasp2025}, which differs from VolCoDiff in three points: 1. No hand branch; 2. The object encoder is PointNet++~\cite{qi2017pointnetplusplus}, and the input are the point positions and normal vectors; 3. The input contact for each point is the vector consisting of the contact and CSE $(c\mid \mathbf{e})$ instead of the latent vector. This model is trained by the standard diffusion loss $\mathcal{L}_{diff}$, building up by the contact value loss $\mathcal{L}_{c}$ and the cse value loss weighted by contact like in VolumeVAE $\mathcal{L}_{cse}$:

\begin{equation}
    \mathcal{L}_{diff} = \mathcal{L}_c + \lambda_{cse}\mathcal{L}_{cse}.
\end{equation}
where $\lambda_{cse}=0.05$ for more focus on the contact.

\subsection{Initialization for Reconstruction From Contact}
\label{app:init}

The methods in Tab.~\ref{tab:gt_contact_recon} obtain the final result by refining an initial pose. This initialization is provided based on the correspondence representations. Note that all initial poses are the average pose, the difference is only in the way to obtain the global 6D pose.

If the contact representation does not encode correspondence \cite{gradyContactOptOptimizingContact2021}), then the initial 6D pose is simply set to 0.

If the correspondence is represented by hand part \cite{liuContactGenGenerativeContact2023}, the initialization can be regarded as aligning the part centers on the rigid hand to those calculated from the contact map. The part centers of the hand are represented by anchors from \cite{yangCPFLearningContact2021}, with picking one anchor to stand for each part. The part centers from the contact is the average of all point positions weighted by the contact likelihood. Denote the hand part centers as $\{\mathbf{p}_i\}_{i=1}^{16}$ and the object part centers as $\{\mathbf{q}_i\}_{i=1}^{16}$, then the initial orientation $R_0$ and translation $\mathbf{t}_0$ are calculated by point cloud registration.

Further, if the correspondence is represented by CSE, then we can calculate the corresponding point on the hand surface for each point in VolCo, by weighted average of relative hand vertices using weights in Eq.~\eqref{eq:recon}. Then $R_0$ and $t_0$ are also calculated by point registration using such point pairs.

\subsection{Contact Area Metric Definition}
\label{app:contact_area}


\textbf{Contact Area (CA)} is proposed to measure the surface area of the object mesh that is in contact
with the hand.
Formally, an object vertex $v_i$ is considered in contact if its distance to the nearest point
on the hand mesh is below a threshold $\delta$ and the object surface normal at $v_i$ opposes
the hand surface normal at the corresponding closest point, \ie
\begin{equation}
    \mathcal{C} = \left\{ v_i \in \mathcal{V}_\text{obj} \mid d(v_i, \mathcal{M}_\text{hand}) < \delta \;\wedge\; \mathbf{n}^{\text{obj}}_i \cdot \mathbf{n}^{\text{hand}}_i < 0 \right\},
\end{equation}
where $\mathcal{V}_\text{obj}$ denotes the object vertices, $d(\cdot,\cdot)$ is the closest-point
distance to the hand mesh $\mathcal{M}_\text{hand}$, and $\mathbf{n}^{\text{obj}}_i$,
$\mathbf{n}^{\text{hand}}_i$ are the surface normals at $v_i$ and its closest hand point,
respectively.
The contact area is then computed as the summed area of all object faces that contain at least
one contact vertex:
\begin{equation}
    \text{CA} = \sum_{f \in \mathcal{F}_\text{obj}} \mathbf{1}\!\left[ f \cap \mathcal{C} \neq \emptyset \right] \cdot A(f),
\end{equation}
where $\mathcal{F}_\text{obj}$ is the set of object faces and $A(f)$ denotes the area of face $f$.
A larger CA indicates more extensive hand-object contact, which is generally desirable for
stable grasps.

\section{Generated Grasp Visualizations}
\label{app:exp_result}

\subsection{Initial Guess and Refinement}

\begin{figure}[t]
    \centering
    \includegraphics[width=0.9\linewidth]{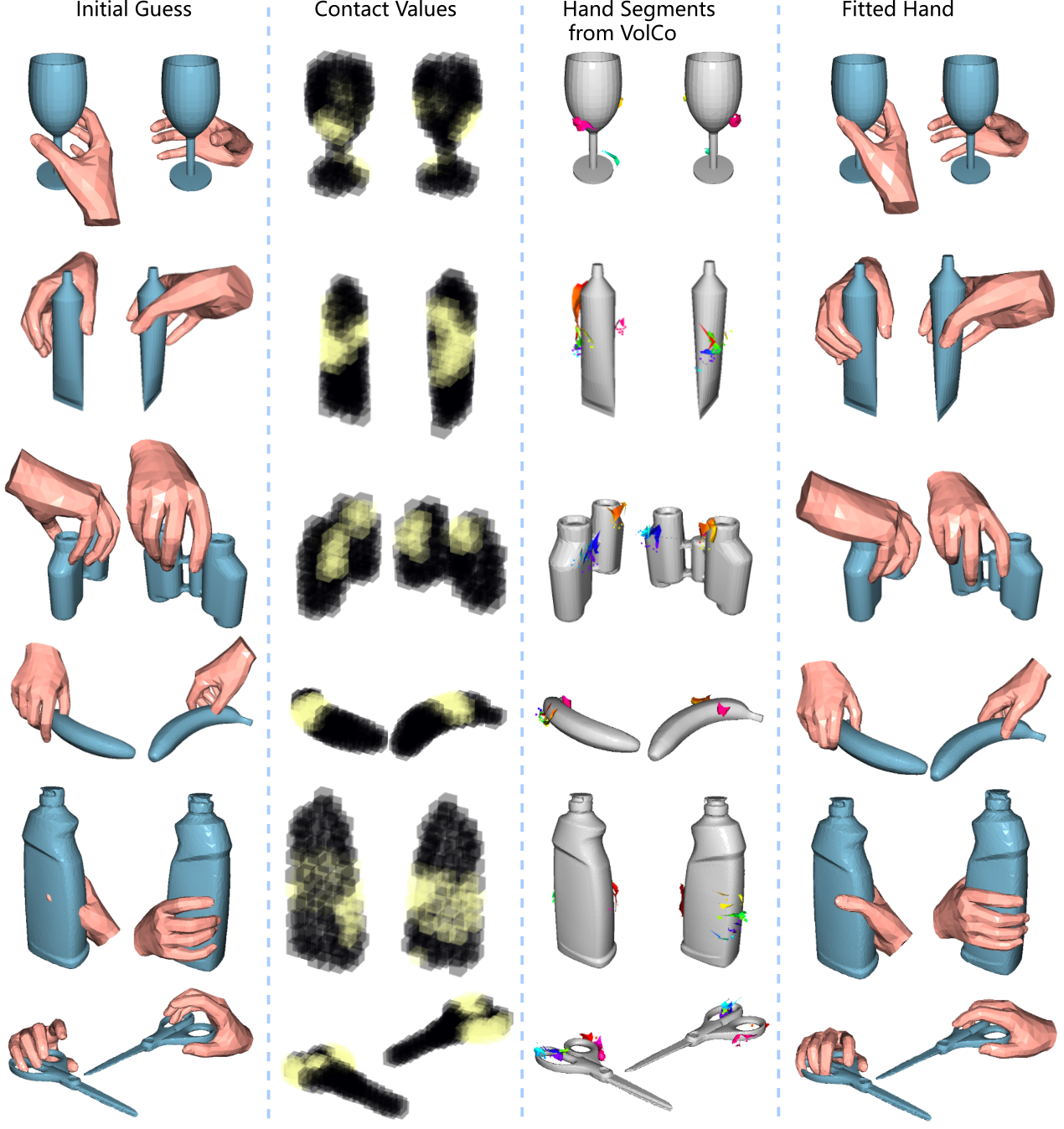}
    \caption{The initial guess, predicted VolCo, and the final grasps of several objects, with each sample shown by two viewpoints.}
    \label{fig:opt_proc}
\end{figure}

To illustrate the process of pose refinement, we show some samples including the initial guess form the hand branch, the contact values shown in volumes, the segments reconstructed from VolCo, and the final fitted hand in Fig.~\ref{fig:opt_proc}. We observe that the initial guess is usually closely related to the predicted VolCo, with only some slight displacement that results in artifacts. With the contact and repulsive losses defined in Eq.~(9), the pose fitting stage efficiently converges to the optimal hand pose consistent with VolCo, for two reasons. First, The dominant loss term $\mathcal{L}_{rec}$ is essentially quadratic w.r.t. the hand vertex locations, yielding a well-behaved optimization landscape without the complexity introduced by contact calculations \cite{gradyContactOptOptimizingContact2021}; Second, the initial guess provides a strong initialization in the vicinity of the optimal pose, facilitating the convergence to the global minimum.

\subsection{Generated Grasps}

In support to the conclusions in the experiments, this section shows more generated samples. These samples are randomly selected for the highest representative. Fig.~\ref{fig:grab_eg} shows GRAB samples Fig.~\ref{fig:ho3d_eg} shows HO3D samples, with each row showing the 4 viewpoints of a sample. The figures demonstrates the diversity and high probability of plausible generation. 

\begin{figure}[p]
    \centering
    \subfloat[Wineglass]{
        \includegraphics[width=.47\textwidth]{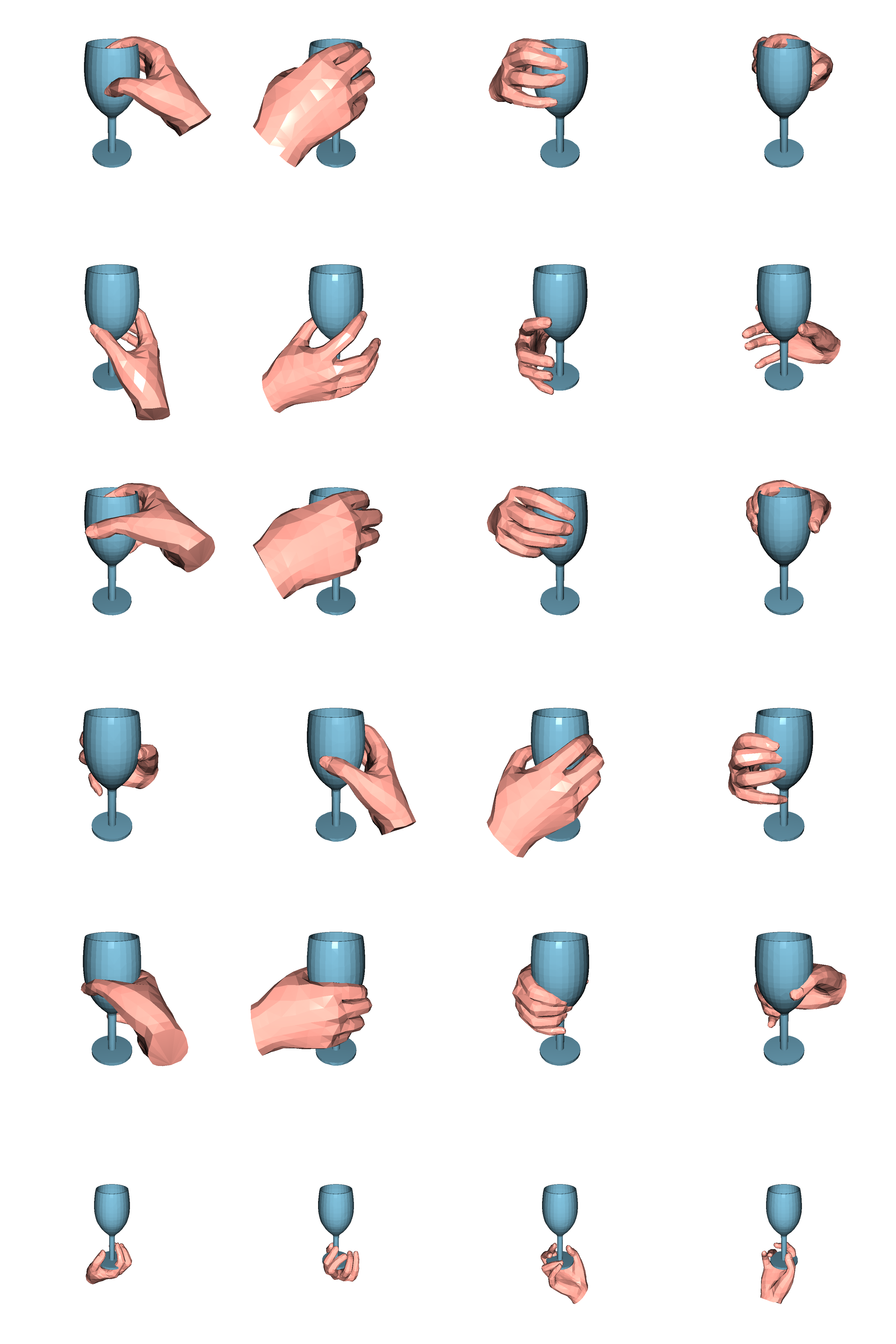}
    }\hfill
    \subfloat[Frying Pan]{
        \includegraphics[width=.47\textwidth]{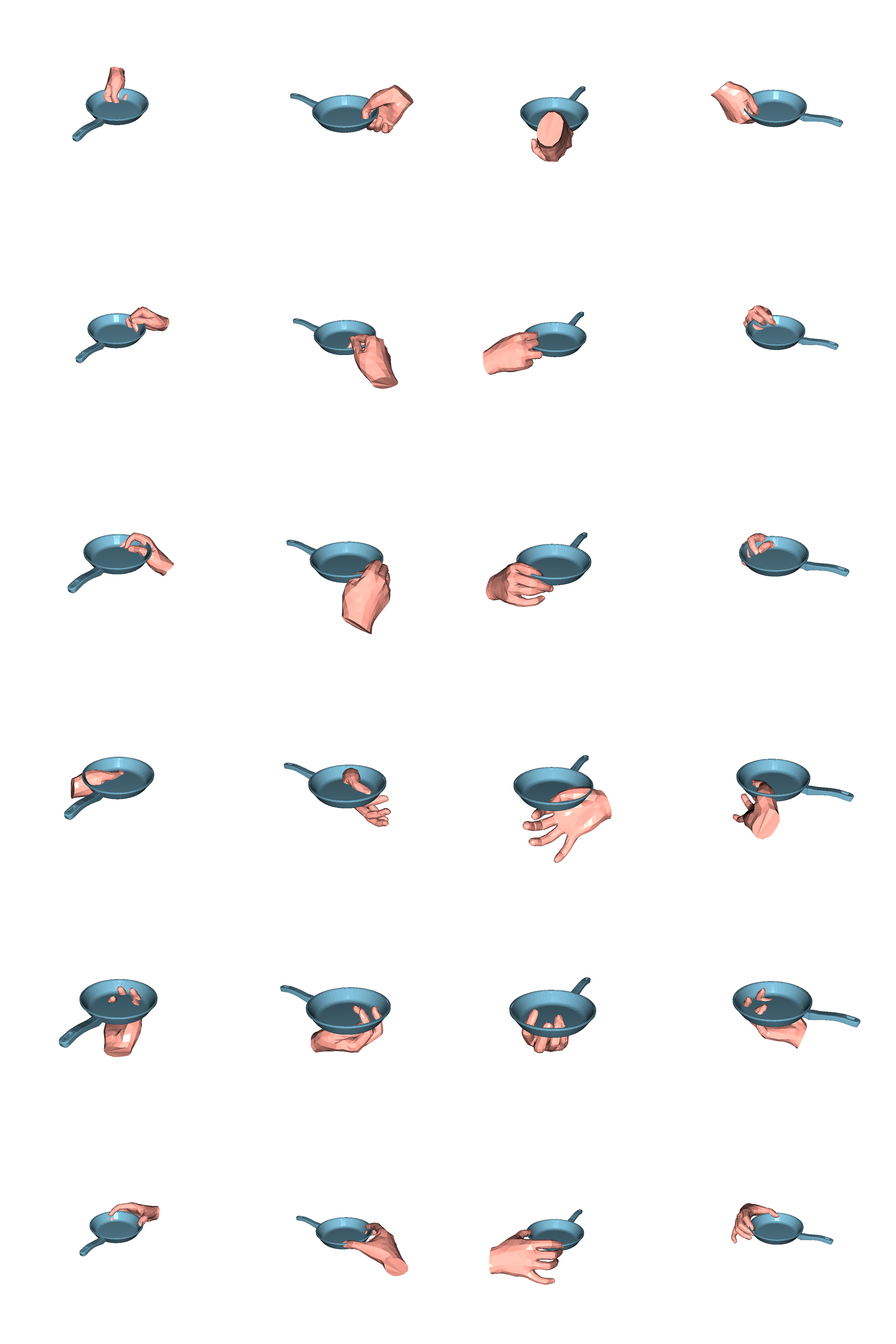}
    }\\
    \subfloat[Mug]{
        \includegraphics[width=.47\textwidth]{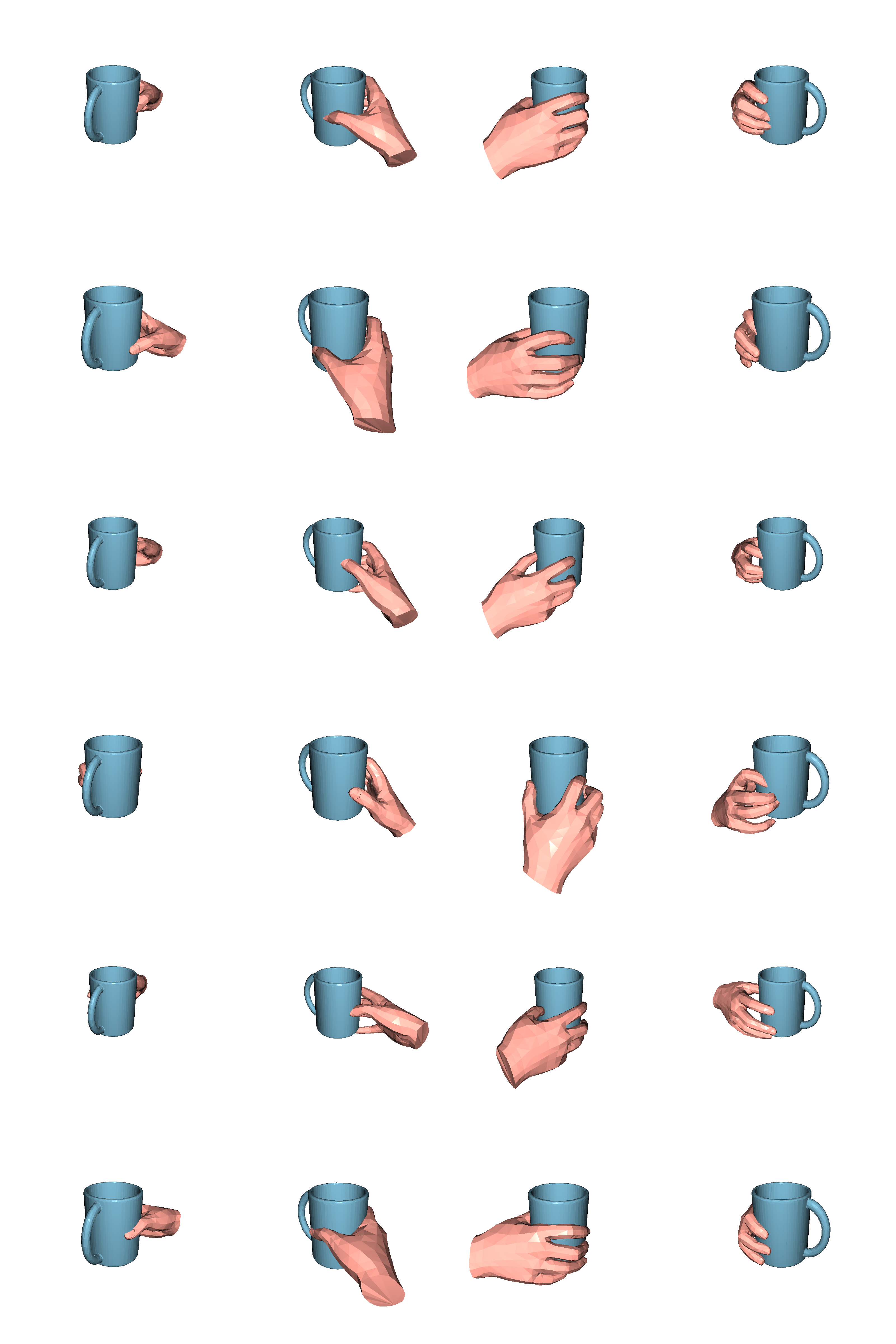}
    }\hfill
    \subfloat[Toothpaste]{
        \includegraphics[width=.47\textwidth]{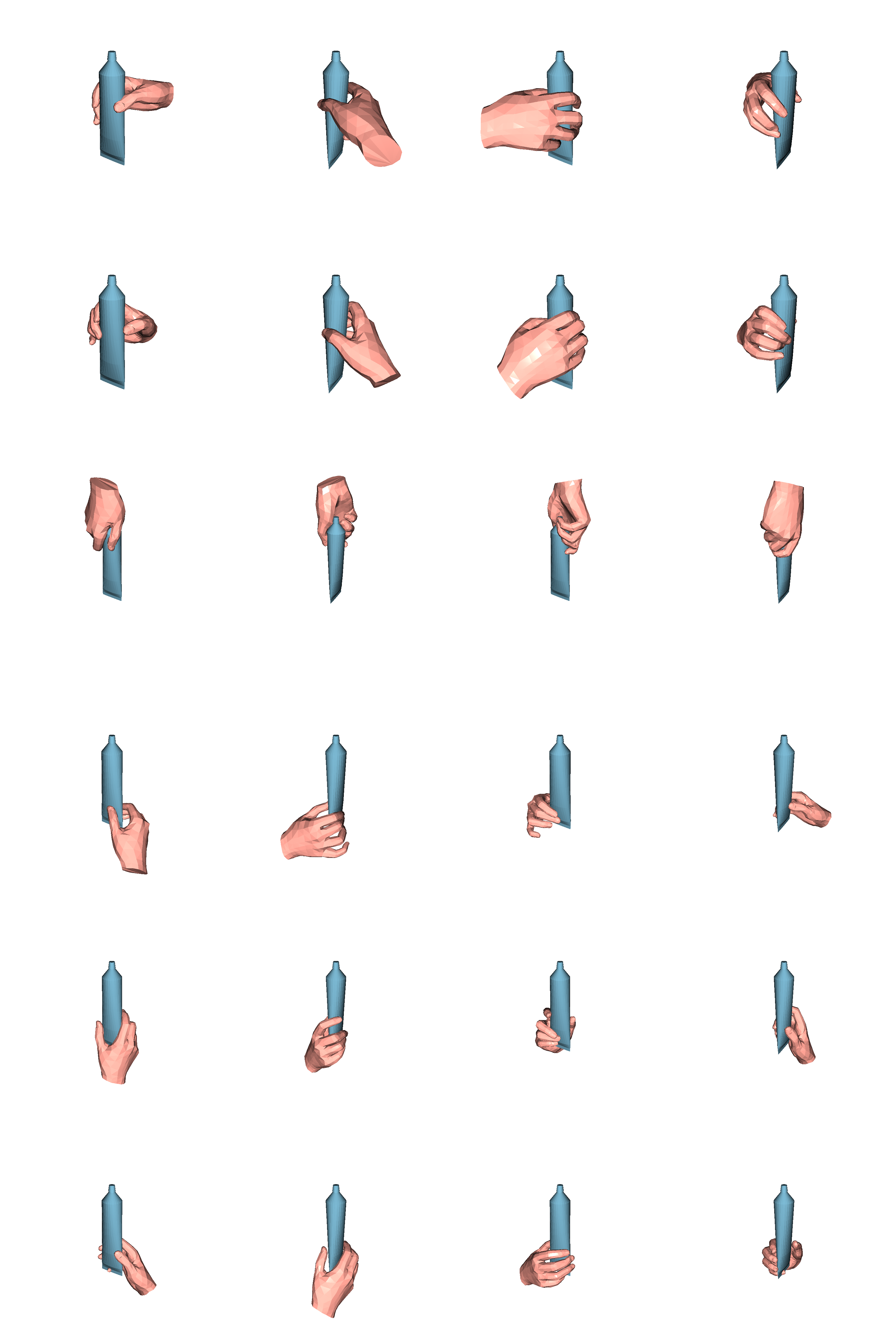}
    }
    \caption{Qualitative results on GRAB dataset (2/3)}
    \label{fig:grab_eg}
\end{figure}

\begin{figure}[t]
    \ContinuedFloat
    \centering
    \subfloat[Camera]{
        \includegraphics[width=.45\textwidth]{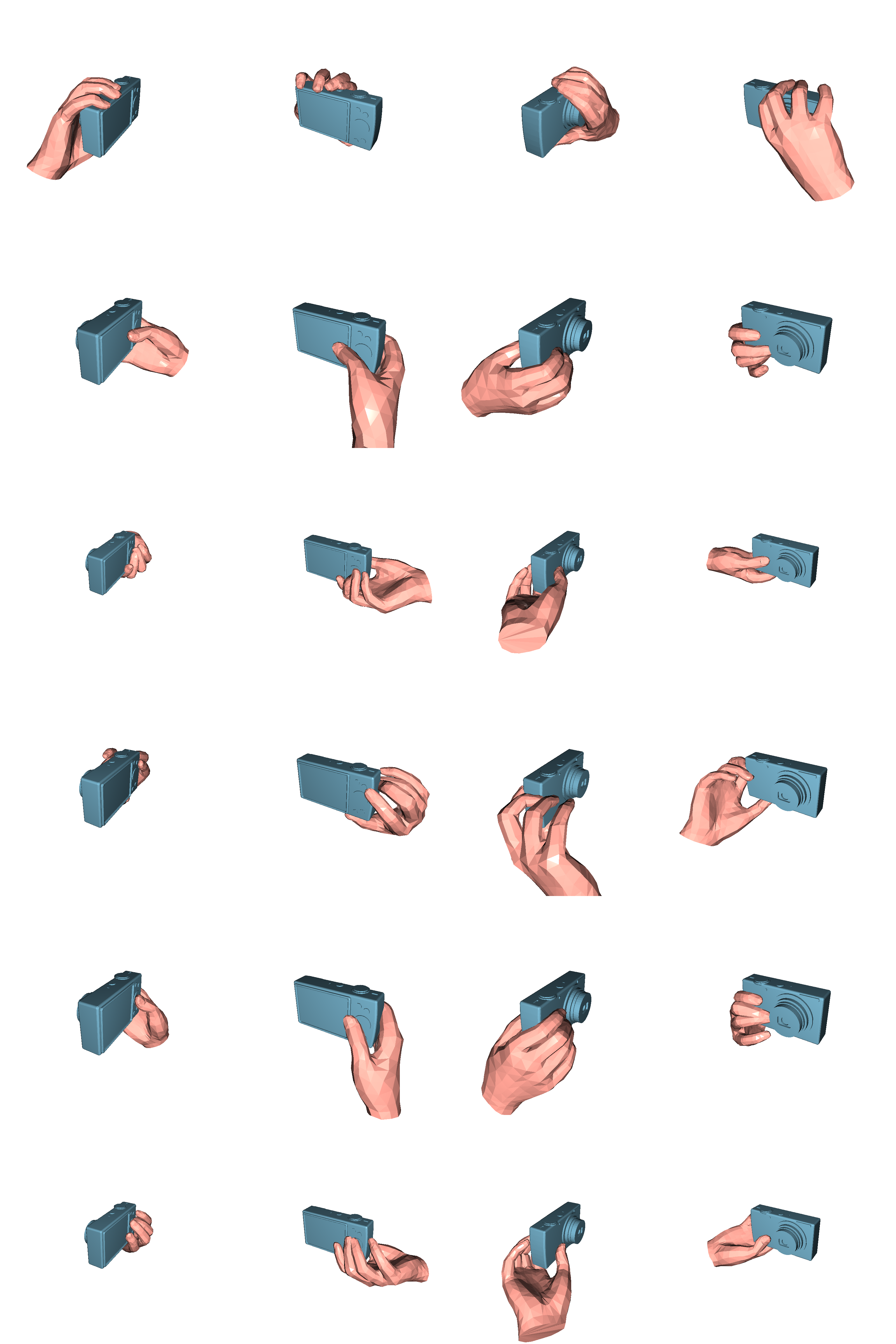}
    }\hfill
    \subfloat[Binoculars]{
        \includegraphics[width=.45\textwidth]{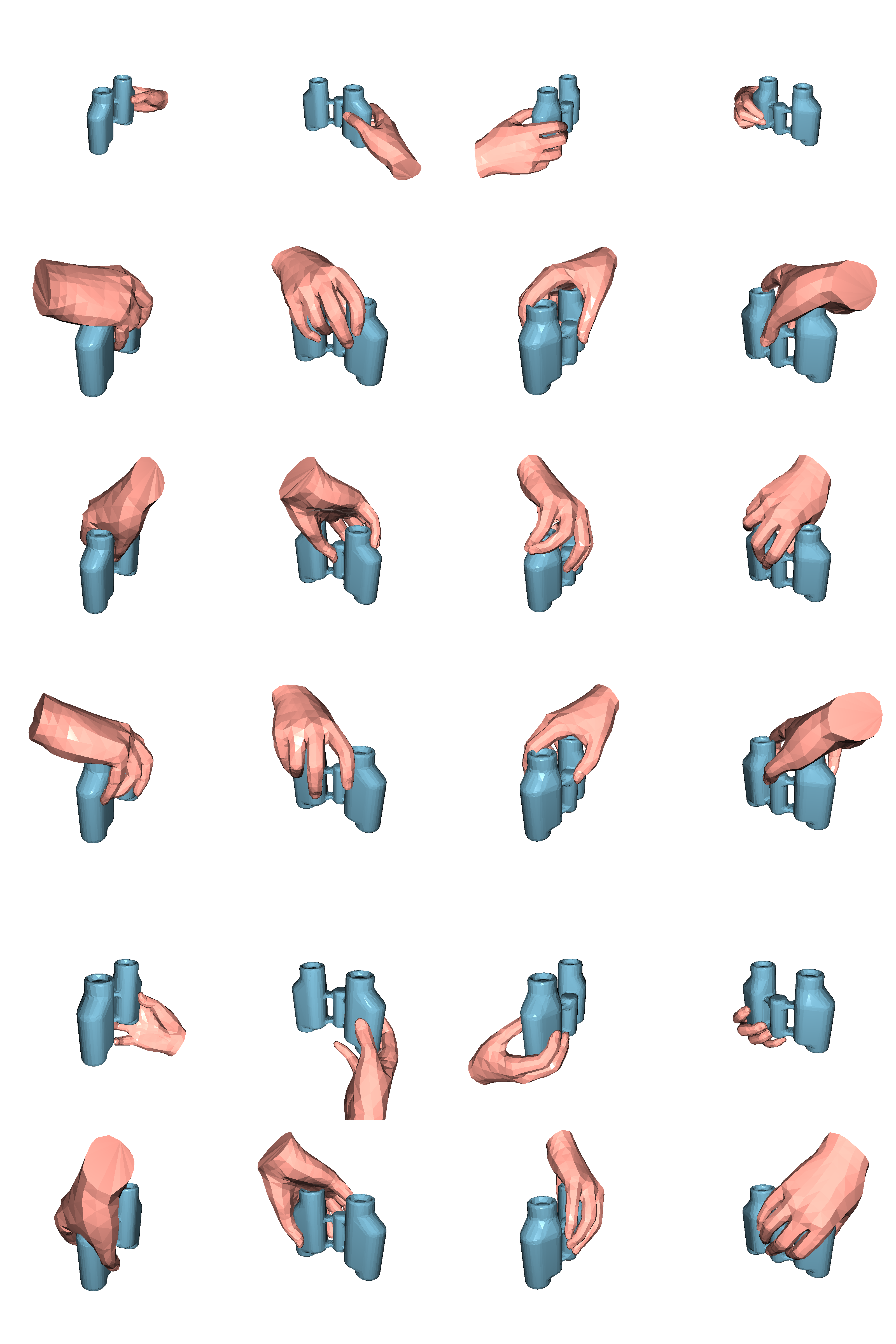}
    }
    \caption{Qualitative results on GRAB dataset (3/3).}
    \label{fig:grab_eg1}
\end{figure}

\begin{figure}[t]
    \centering
    \subfloat[Banana]{
        \includegraphics[width=.45\textwidth]{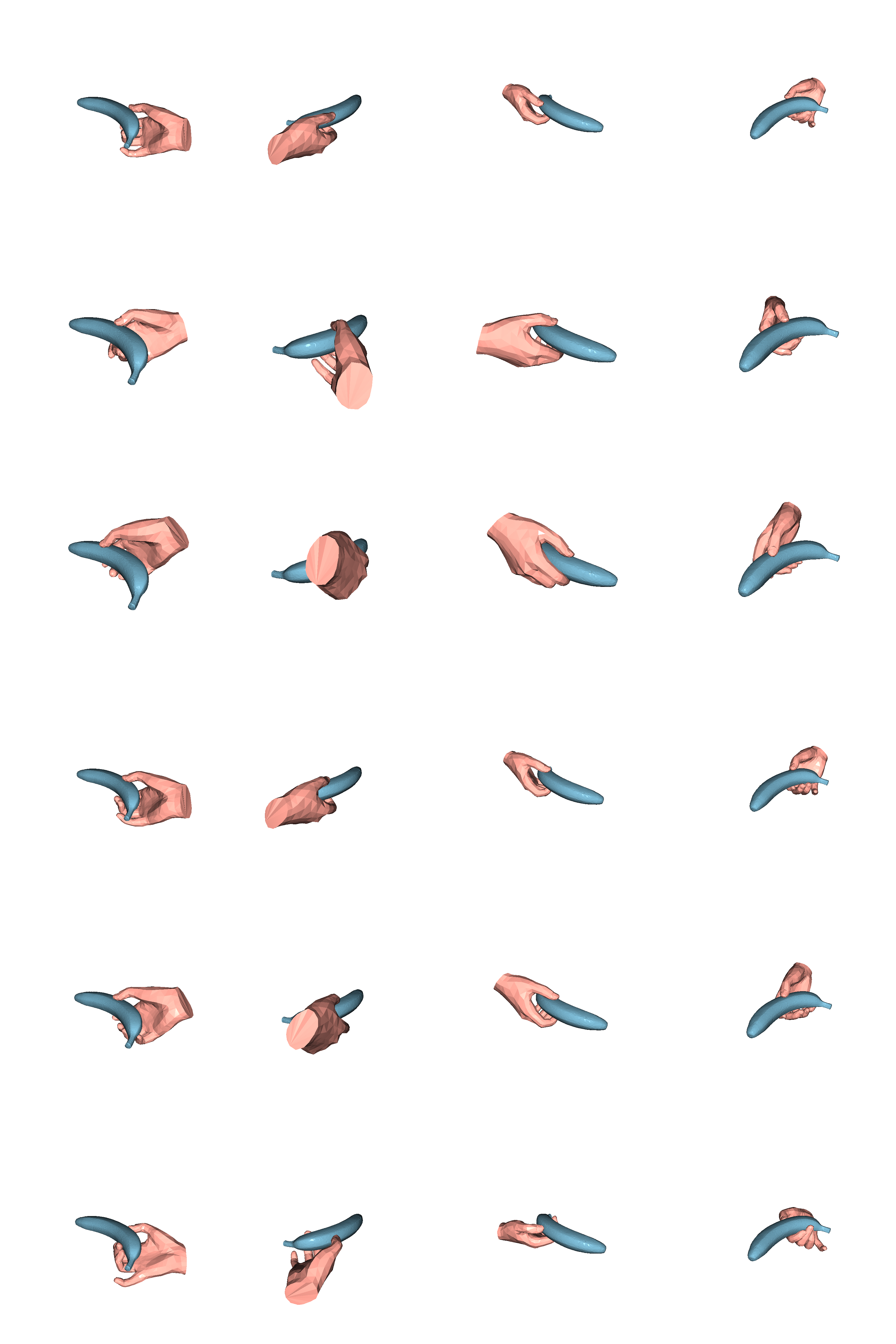}
    }\hfill
    \subfloat[Bleach Cleanser]{
        \includegraphics[width=.45\textwidth]{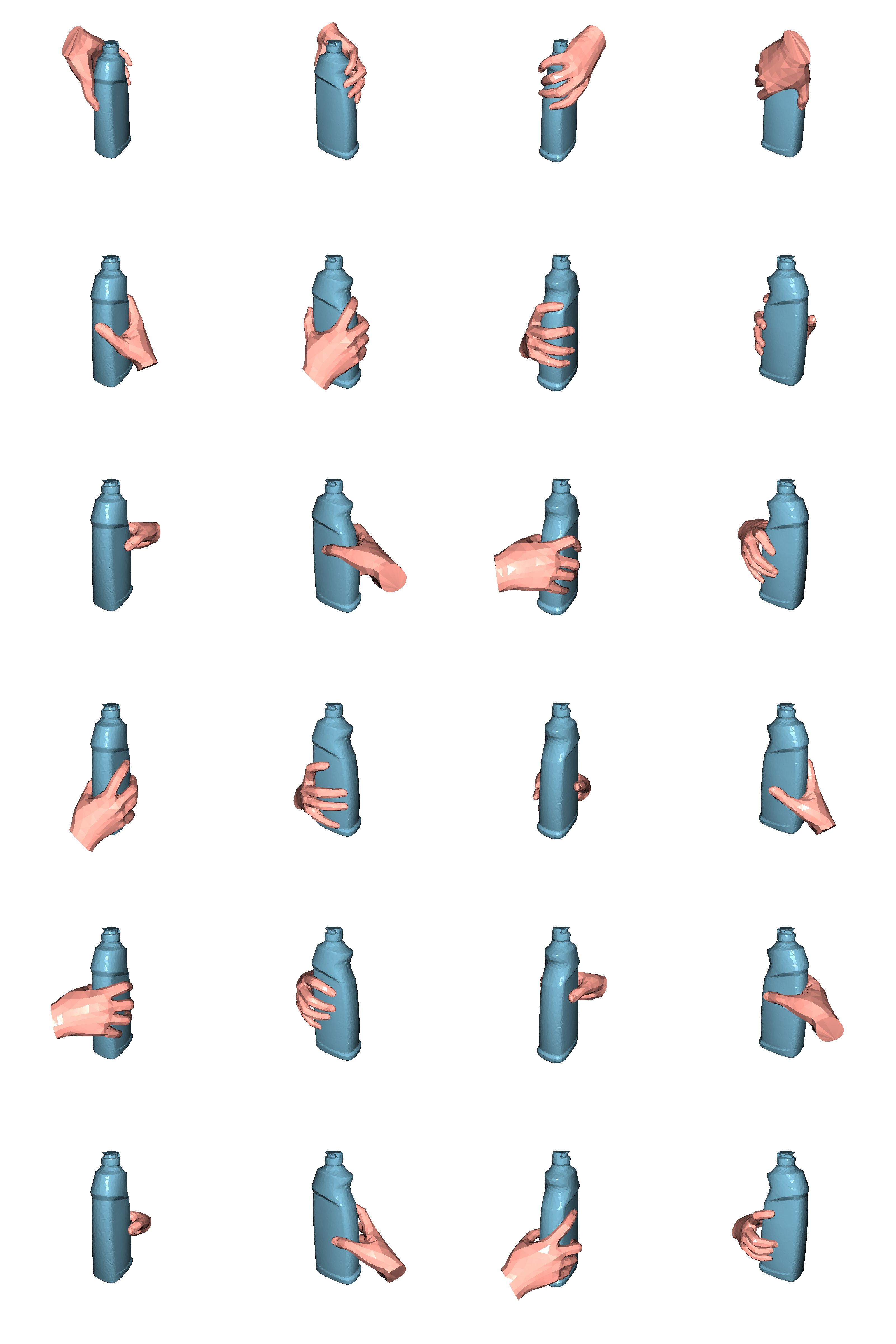}
    }
    \caption{Qualitative results on out-of-domain HO3D dataset (1/5).}
    \label{fig:ho3d_eg}
\end{figure}

\begin{figure}[p]
    \ContinuedFloat
    \centering
    \subfloat[Cracker Box]{
        \includegraphics[width=.47\textwidth]{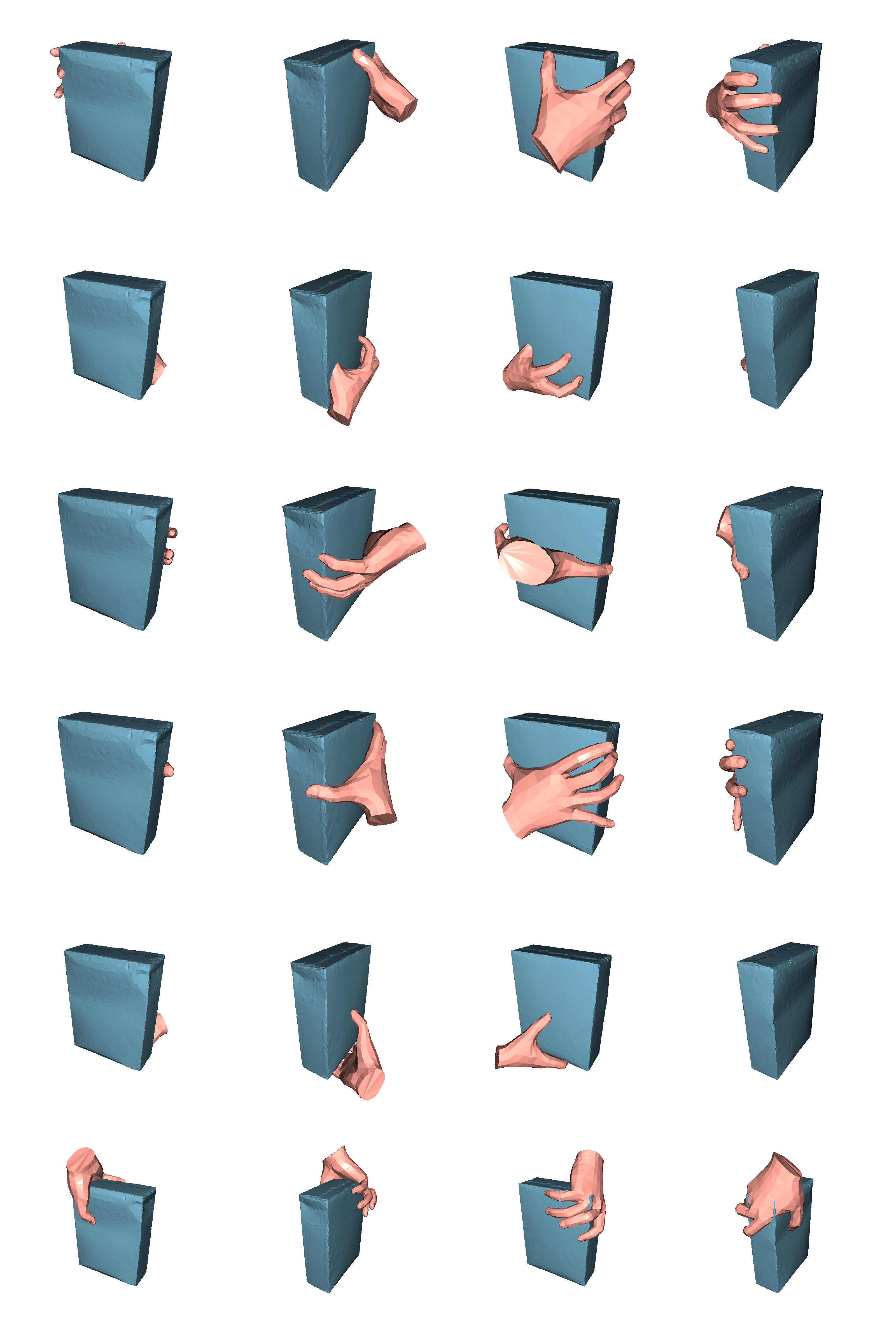}
    }\hfill
    \subfloat[Power Drill]{
        \includegraphics[width=.47\textwidth]{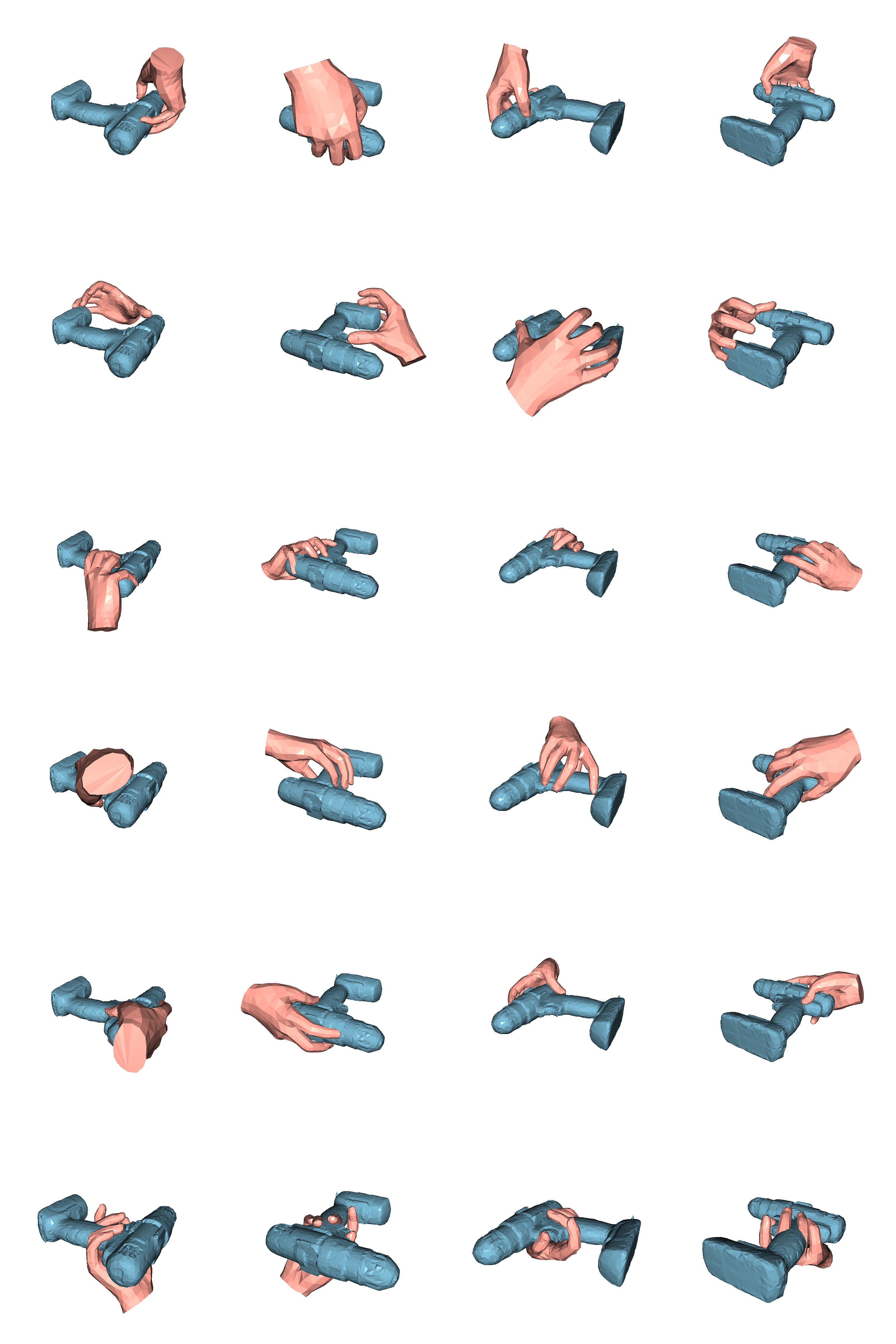}
    }\\
    \subfloat[Mug]{
        \includegraphics[width=.47\textwidth]{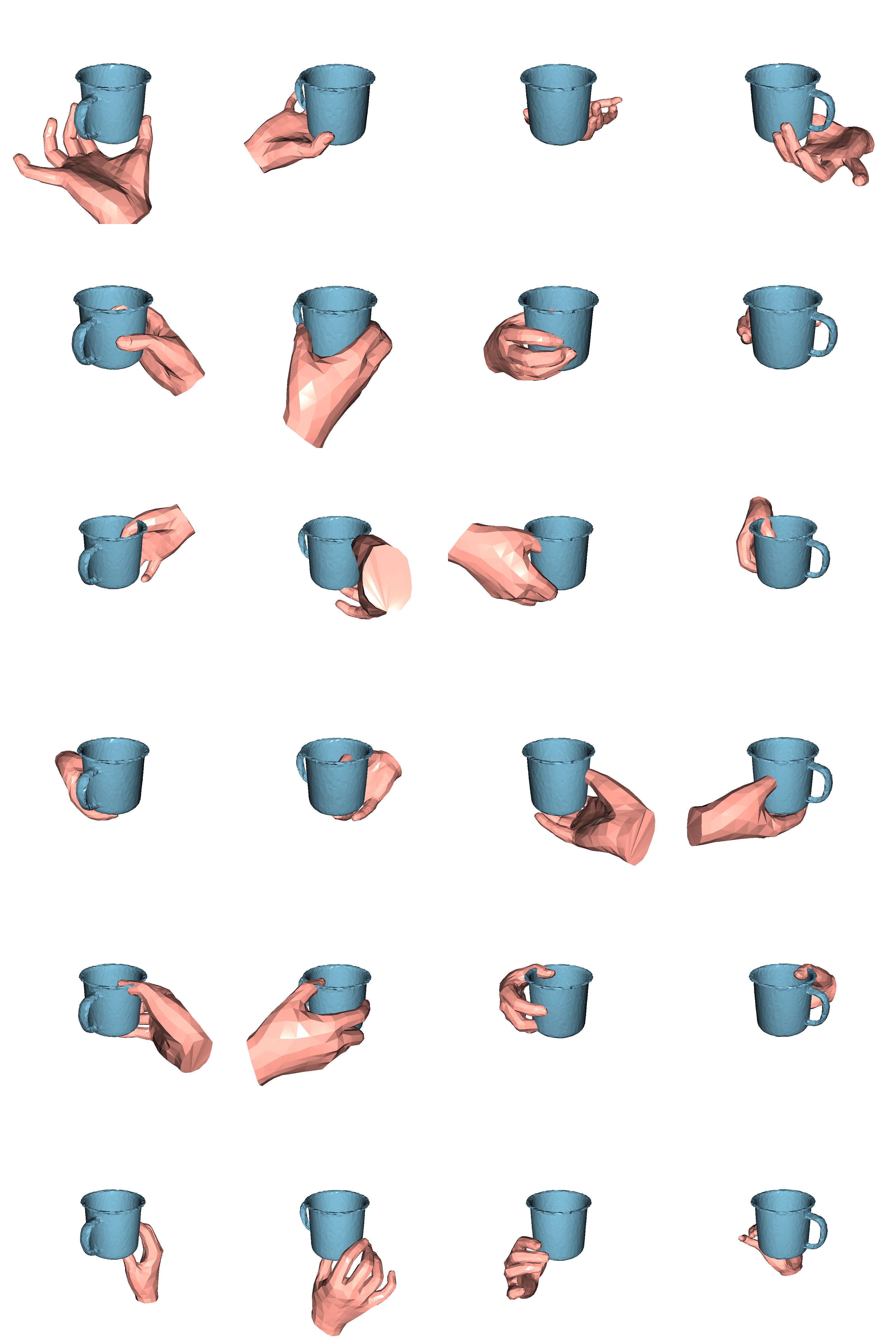}
    }\hfill
    \subfloat[Mustard Bottle]{
        \includegraphics[width=.47\textwidth]{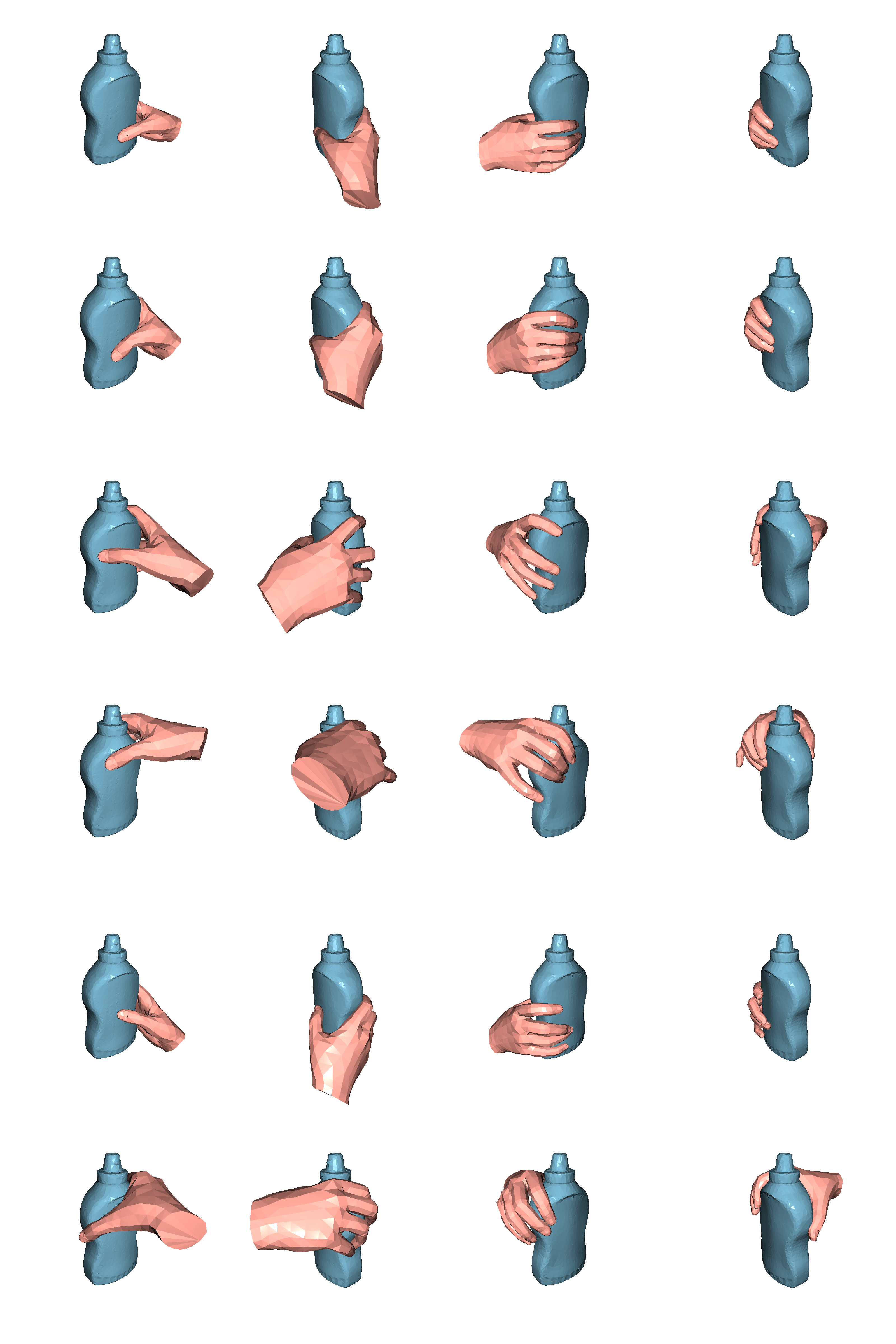}
    }
    \caption{Qualitative results on out-of-domain HO3D dataset (3/5).}
    \label{fig:ho3d_eg1}
\end{figure}

\begin{figure}[p]
    \ContinuedFloat
    \centering
    \subfloat[Pitcher Base]{
        \includegraphics[width=.47\textwidth]{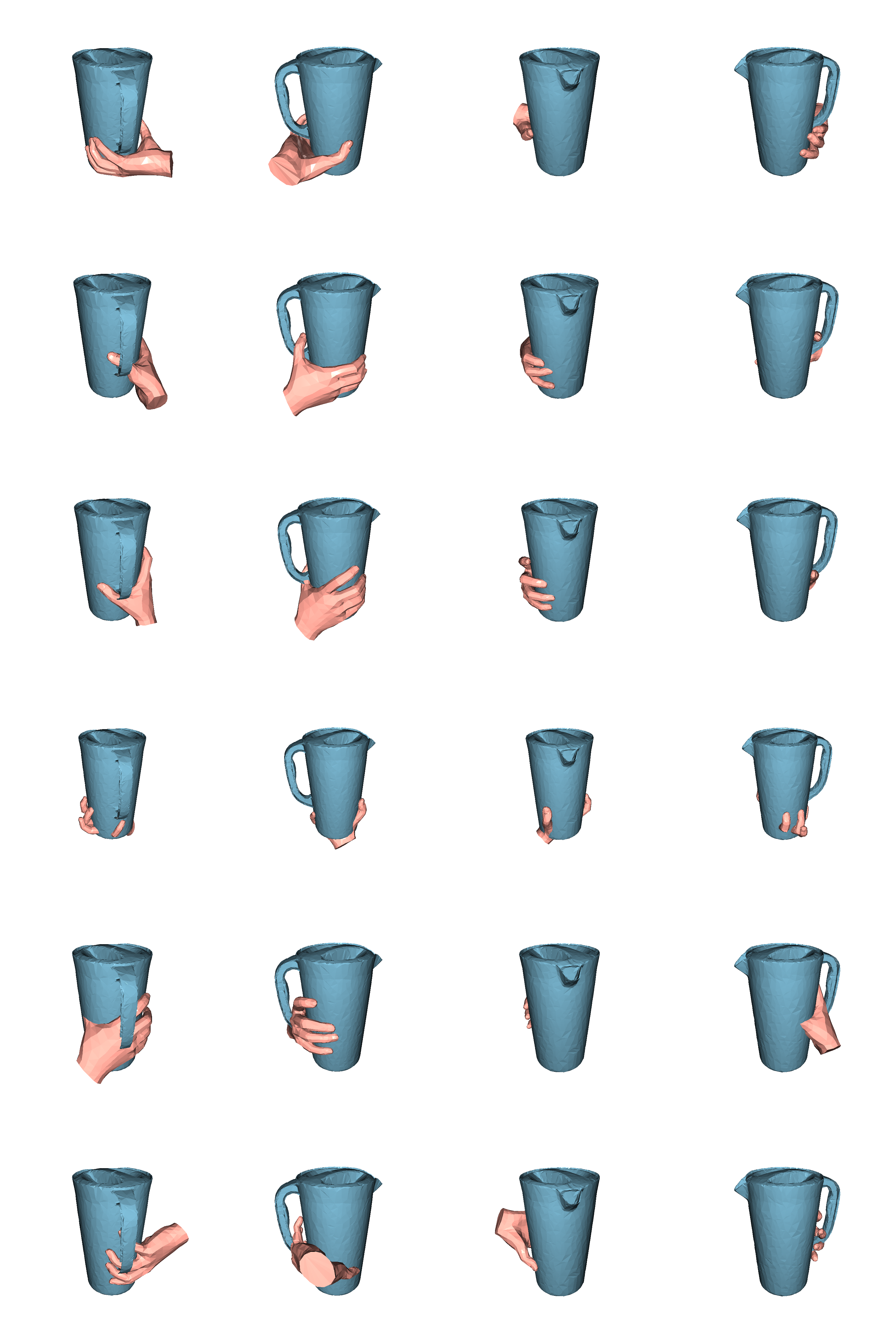}
    }\hfill
    \subfloat[Potted Meat Can]{
        \includegraphics[width=.47\textwidth]{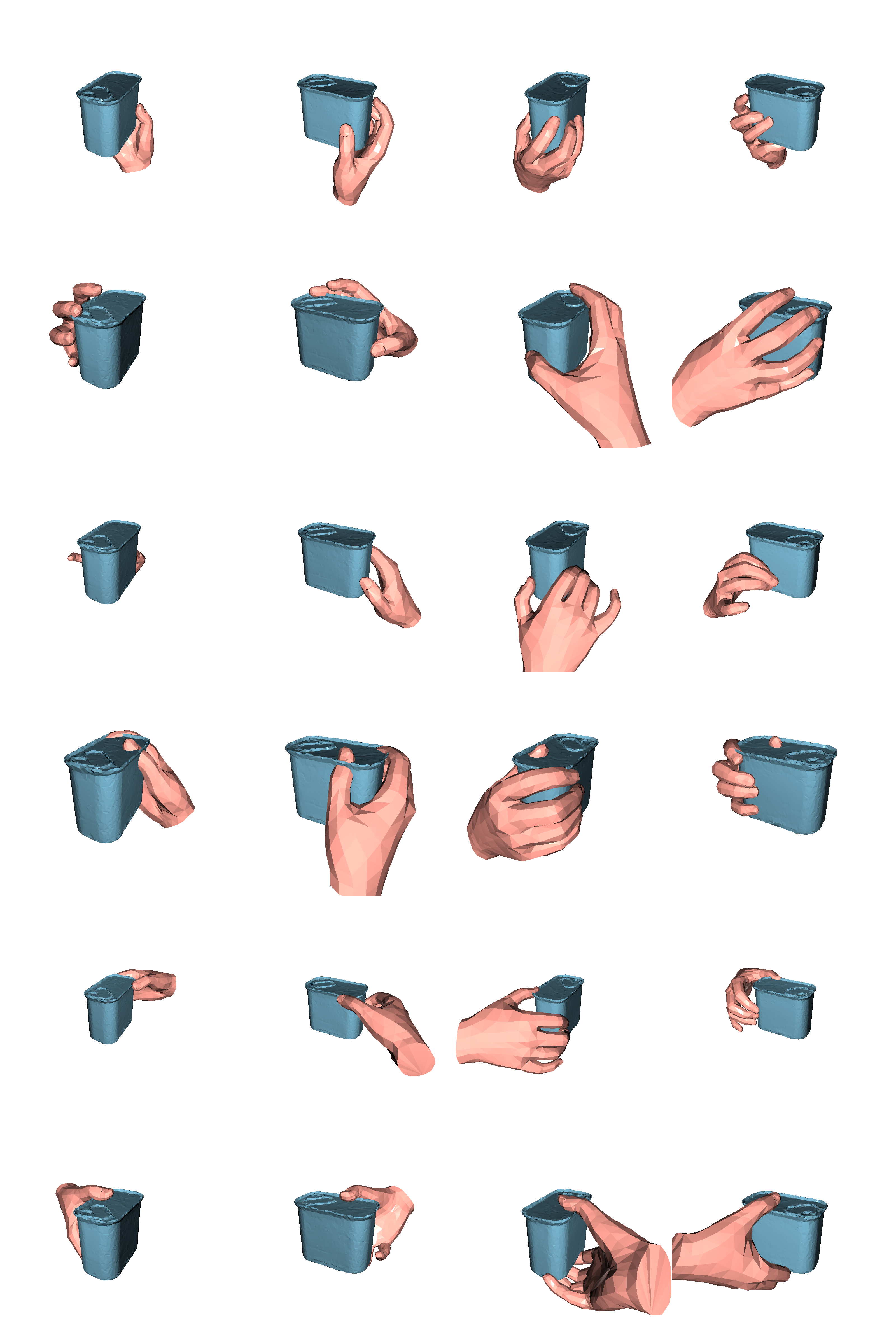}
    }\\
    \subfloat[Scissors]{
        \includegraphics[width=.47\textwidth]{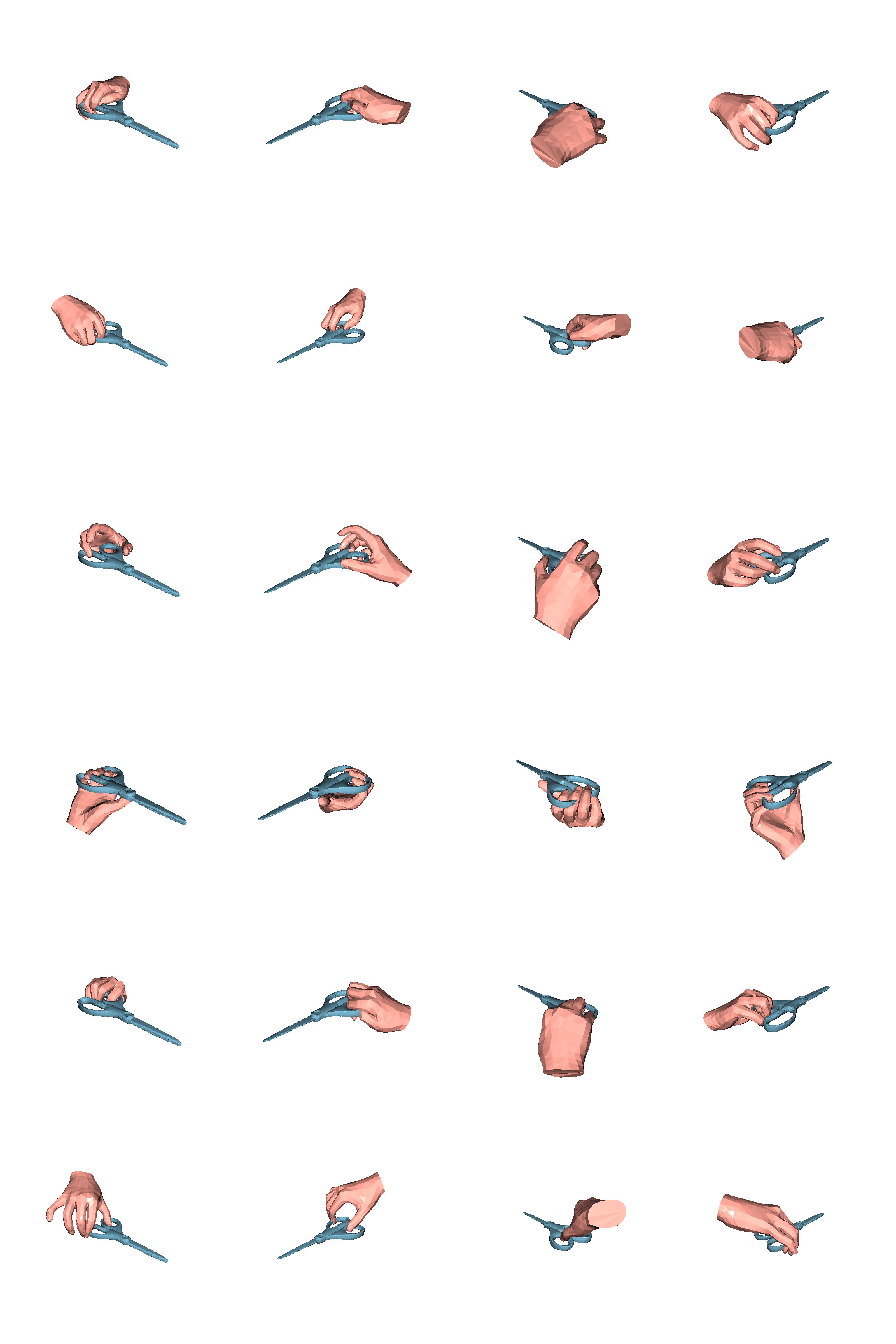}
    }\hfill
    \subfloat[Sugar Box]{
        \includegraphics[width=.47\textwidth]{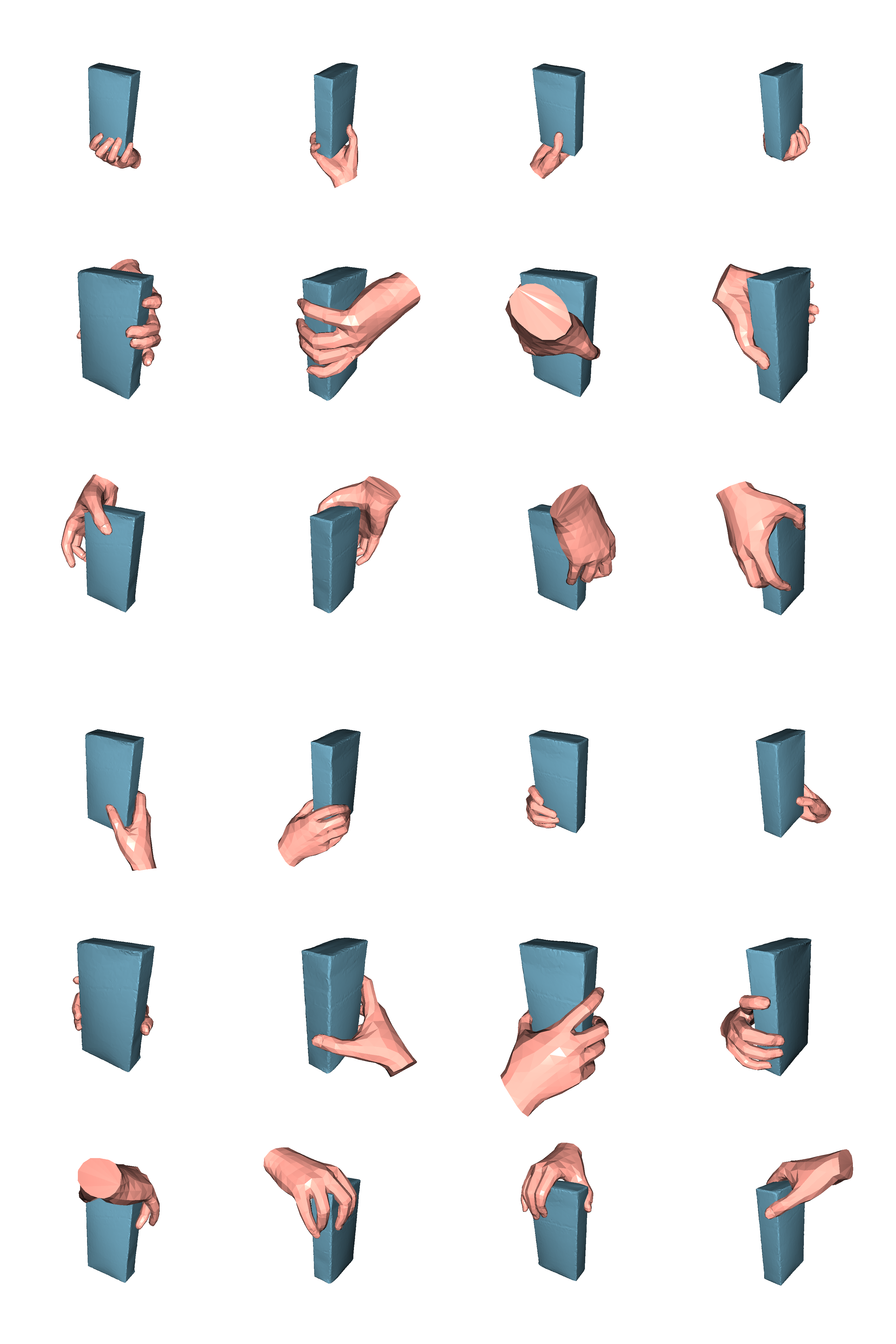}
    }
    \caption{Qualitative results on out-of-domain HO3D dataset (5/5).}
    \label{fig:ho3d_eg2}
\end{figure}




\newpage
\section*{NeurIPS Paper Checklist}

\begin{enumerate}

\item {\bf Claims}
    \item[] Question: Do the main claims made in the abstract and introduction accurately reflect the paper's contributions and scope?
    \item[] Answer: \answerYes{} 
    \item[] Justification: The main contribution of this paper is a novel contact representation based on local volumes and a grasp generation framework based on it. The new method can generate tighter grasps with much less severe penetrations. This is clearly introduced in abstract and introduction.

\item {\bf Limitations}
    \item[] Question: Does the paper discuss the limitations of the work performed by the authors?
    \item[] Answer: \answerYes{} 
    \item[] Justification: The limitations are introduced in the last section.

\item {\bf Theory assumptions and proofs}
    \item[] Question: For each theoretical result, does the paper provide the full set of assumptions and a complete (and correct) proof?
    \item[] Answer: \answerNA{} 
    \item[] Justification: The paper does not include theoretical results.

    \item {\bf Experimental result reproducibility}
    \item[] Question: Does the paper fully disclose all the information needed to reproduce the main experimental results of the paper to the extent that it affects the main claims and/or conclusions of the paper (regardless of whether the code and data are provided or not)?
    \item[] Answer: \answerYes{} 
    \item[] Justification: All hyperparameters we use in the training and testing are given in the paper. The model architecture is briefly introduced in the paper and described in detail in Appendix.~\ref{app:model}.

\item {\bf Open access to data and code}
    \item[] Question: Does the paper provide open access to the data and code, with sufficient instructions to faithfully reproduce the main experimental results, as described in supplemental material?
    \item[] Answer: \answerNo{} 
    \item[] Justification: The code will be released upon accepted.

\item {\bf Experimental setting/details}
    \item[] Question: Does the paper specify all the training and test details (e.g., data splits, hyperparameters, how they were chosen, type of optimizer) necessary to understand the results?
    \item[] Answer: \answerYes{} 
    \item[] Justification: The hyperparameters are given in Methodology section, with brief motivations on choosing these parameters. Mostly, the hyperparameters are selected following previous works. 

\item {\bf Experiment statistical significance}
    \item[] Question: Does the paper report error bars suitably and correctly defined or other appropriate information about the statistical significance of the experiments?
    \item[] Answer: \answerYes{} 
    \item[] Justification: The results reported in the main tables Tab.~\ref{tab:quant_result_grab} and \ref{tab:quant_result_ho3d} are accompanied with error bars indicating the standard deviation (1-sigma).

\item {\bf Experiments compute resources}
    \item[] Question: For each experiment, does the paper provide sufficient information on the computer resources (type of compute workers, memory, time of execution) needed to reproduce the experiments?
    \item[] Answer: \answerYes{} 
    \item[] Justification: The detailed computational sources are described in Appendix.
    
\item {\bf Code of ethics}
    \item[] Question: Does the research conducted in the paper conform, in every respect, with the NeurIPS Code of Ethics \url{https://neurips.cc/public/EthicsGuidelines}?
    \item[] Answer: \answerYes{} 
    \item[] Justification: This paper utilizes only the public datasets for experiments, which are open-access for research purposes. These are no self-collected human data in our experiments.

\item {\bf Broader impacts}
    \item[] Question: Does the paper discuss both potential positive societal impacts and negative societal impacts of the work performed?
    \item[] Answer: \answerYes{} 
    \item[] Justification: The social impacts are discussed in the last section.
    
\item {\bf Safeguards}
    \item[] Question: Does the paper describe safeguards that have been put in place for responsible release of data or models that have a high risk for misuse (e.g., pre-trained language models, image generators, or scraped datasets)?
    \item[] Answer: \answerNA{} 
    \item[] Justification: This paper proposes to generate the human grasp, which aims at studying the general interactive patterns and can hardly be used to harm public safety. Though it can foster the deployment of dexterous robots that might result in harmful context, this is more related with robotic breakthroughs and are out of scope of this paper.

\item {\bf Licenses for existing assets}
    \item[] Question: Are the creators or original owners of assets (e.g., code, data, models), used in the paper, properly credited and are the license and terms of use explicitly mentioned and properly respected?
    \item[] Answer: \answerYes{} 
    \item[] Justification: The public datasets are properly cited and we use them following their terms of use.

\item {\bf New assets}
    \item[] Question: Are new assets introduced in the paper well documented and is the documentation provided alongside the assets?
    \item[] Answer: \answerNA{} 
    \item[] Justification: This paper does not contribute new assets.

\item {\bf Crowdsourcing and research with human subjects}
    \item[] Question: For crowdsourcing experiments and research with human subjects, does the paper include the full text of instructions given to participants and screenshots, if applicable, as well as details about compensation (if any)? 
    \item[] Answer: \answerNA{} 
    \item[] Justification: There is no user study in this paper.

\item {\bf Institutional review board (IRB) approvals or equivalent for research with human subjects}
    \item[] Question: Does the paper describe potential risks incurred by study participants, whether such risks were disclosed to the subjects, and whether Institutional Review Board (IRB) approvals (or an equivalent approval/review based on the requirements of your country or institution) were obtained?
    \item[] Answer: \answerNA{} 
    \item[] Justification: This paper does not include crowdsourcing or research with human subjects.

\item {\bf Declaration of LLM usage}
    \item[] Question: Does the paper describe the usage of LLMs if it is an important, original, or non-standard component of the core methods in this research? Note that if the LLM is used only for writing, editing, or formatting purposes and does \emph{not} impact the core methodology, scientific rigor, or originality of the research, declaration is not required.
    \item[] Answer: \answerNA{} 
    \item[] Justification: The main framework of the method is based on latent diffusion model, not LLM.

\end{enumerate}

\end{document}